\documentclass[runningheads]{llncs}
\usepackage{graphicx}
\usepackage{hyperref}
\usepackage{appendix}

\usepackage{subcaption}
\usepackage[labelformat=parens,labelsep=quad,skip=3pt]{caption}
\usepackage{multirow}
\usepackage{arydshln}
\usepackage[usenames, dvipsnames, table]{xcolor}

\usepackage[ruled,noend]{algorithm2e}

\usepackage{array}
\newcolumntype{x}[1]{>{\centering\let\newline\\\arraybackslash\hspace{0pt}}p{#1}}

\usepackage{amsmath}
\DeclareMathOperator*{\argmax}{argmax}

\usepackage{wrapfig}
\makeatletter 
\def\WFfill{\par 
    \ifx\parshape\WF@fudgeparshape 
    \nobreak 
    \ifnum\c@WF@wrappedlines>\@ne 
    \advance\c@WF@wrappedlines\m@ne 
    \vskip\c@WF@wrappedlines\baselineskip 
    \global\c@WF@wrappedlines\z@ 
    \fi 
    \allowbreak 
    \WF@finale 
    \fi 
} 
\makeatother 

\begin{document}
\title{Data Poisoning Attacks Against\\Federated Learning Systems}
%
%
\author{Vale Tolpegin \and 
Stacey Truex \and 
Mehmet Emre Gursoy \and 
Ling Liu}
\authorrunning{V. Tolpegin et al.}
%
\institute{Georgia Institute of Technology, Atlanta GA 30332, USA 
\email{\{vtolpegin3,memregursoy,staceytruex\}@gatech.edu, ling.liu@cc.gatech.edu}}
%
\maketitle              

\begin{abstract}
Federated learning (FL) is an emerging paradigm for distributed training of large-scale deep neural networks in which participants' data remains on their own devices with only model updates being shared with a central server. However, the distributed nature of FL gives rise to new threats caused by potentially malicious participants. In this paper, we study targeted data poisoning attacks against FL systems in which a malicious subset of the participants aim to poison the global model by sending model updates derived from mislabeled data. We first demonstrate that such data poisoning attacks can cause substantial drops in classification accuracy and recall, even with a small percentage of malicious participants. We additionally show that the attacks can be targeted, i.e., they have a large negative impact only on classes that are under attack. We also study attack longevity in early/late round training, the impact of malicious participant availability, and the relationships between the two. Finally, we propose a defense strategy that can help identify malicious participants in FL to circumvent poisoning attacks, and demonstrate its effectiveness.

\end{abstract}
\section{Introduction}\label{sec:intro}

\vspace{-10pt}
Machine learning (ML) has become ubiquitous in today's society as a range of industries deploy predictive models into their daily workflows. This environment has not only put a premium on the ML model training and hosting technologies but also on the rich data that companies are collecting about their users to train and inform such models. Companies and users alike are consequently faced with 2 fundamental questions in this reality of ML: (1) How can privacy concerns around such pervasive data collection be moderated without sacrificing the efficacy of ML models? and (2) How can ML models be trusted as accurate predictors?

Federated ML has seen increased adoption in recent years~\cite{hard2018federated,ryffel2018generic,47976} in response to the growing legislative demand to address user privacy~\cite{act1996health,mathews2018california,regulation2016regulation}. Federated learning (FL) allows data to remain at the edge with only model parameters being shared with a central server. Specifically, there is no centralized data curator who collects and verifies an aggregate dataset. Instead, each data holder (participant) is responsible for conducting training on their local data. In regular intervals participants are then send model parameter values to a central parameter server or aggregator where a global model is created through aggregation of the individual updates. A global model can thus be trained over all participants' data without any individual participant needing to share their private raw data.

While FL systems allow participants to keep their raw data local, a significant vulnerability is introduced at the heart of question (2). Consider the scenario wherein a subset of participants are either malicious or have been compromised by some adversary. This can lead to these participants having mislabeled or poisonous samples in their local training data. With no central authority able to validate data, these malicious participants can consequently poison the trained global model. For example, consider Microsoft's AI chat bot Tay. Tay was released on Twitter with the underlying natural language processing model set to learn from the 
Twitter users it interacted with. Thanks to malicious users, Tay was quickly manipulated to learn offensive and racist language~\cite{schlesinger2018let}. 

In this paper, we study the vulnerability of FL systems to malicious participants seeking to poison the globally trained model. We make minimal assumptions on the capability of a malicious FL participant -- each can only manipulate the raw training data on their device. This allows for non-expert malicious participants to achieve poisoning with no knowledge of model type, parameters, and FL process. Under this set of assumptions, label flipping attacks become a feasible strategy to implement data poisoning, attacks which have been shown to be effective against traditional, centralized ML models \cite{biggio2011support,steinhardt2017certified,xiao2012adversarial,xiao2015support}. We investigate their application to FL systems using complex deep neural network models.

We demonstrate our FL poisoning attacks using two popular image classification datasets: CIFAR-10 and Fashion-MNIST. Our results yield several interesting findings. First, we show that attack effectiveness (decrease in model utility) depends on the percentage of malicious users and the attack is effective even when this percentage is small. Second, we show that attacks can be targeted, i.e., they have large negative impact on the subset of classes that are under attack, but have little to no impact on remaining classes. This is desirable for adversaries who wish to poison a subset of classes while not completely corrupting the global model to avoid easy detection. Third, we evaluate the impact of attack timing (poisoning in early or late rounds of FL training) and the impact of malicious participant availability (whether malicious participants can increase their availability and selection rate to increase effectiveness). Motivated by our finding that the global model may still converge accurately after early-round poisoning stops, we conclude that largest poisoning impact can be achieved if malicious users participate in later rounds and with high availability.

Given the highly effective poisoning threat to FL systems, we then propose a defense strategy for the FL aggregator to identify malicious participants using their model updates. Our defense is based on the insight that updates sent from malicious participants have unique characteristics compared to honest participants' updates. Our defense extracts relevant parameters from the high-dimensional update vectors and applies PCA for dimensionality reduction. Results on CIFAR-10 and Fashion-MNIST across varying malicious participant rates (2-20\%) show that the aggregator can obtain clear separation between malicious and honest participants' respective updates using our defense strategy. This enables the FL aggregator to identify and block malicious participants. 

The rest of this paper is organized as follows. In Section \ref{sec:attack_eval}, we introduce the FL setting, threat model, attack strategy, and attack evaluation metrics. In Section \ref{sec:experiments}, we demonstrate the effectiveness of FL poisoning attacks and analyze their impact with respect to malicious participant percentage, choice of classes under attack, attack timing, and malicious participant availability. In Section \ref{sec:defense}, we describe and empirically demonstrate our defense strategy. We discuss related work in Section \ref{sec:related_work} and conclude in Section \ref{sec:conclusion}. Our source code is available\footnote{\url{https://github.com/git-disl/DataPoisoning_FL}}. 

\vspace{-12pt}
\section{Preliminaries and Attack Formulation}\label{sec:attack_eval}
\vspace{-6pt}
\subsection{Federated Machine Learning}\label{subsec:fl}
\vspace{-6pt}

FL systems allow global model training without the sharing of raw private data. Instead, individual participants only share model parameter updates. Consider a deep neural network (DNN) model. DNNs consist of multiple layers of nodes where each node is a basic functional unit with a corresponding set of parameters. Nodes receive input from the immediately preceding layer and send output to the following layer; with the first layer nodes receiving input from the training data and the final layer nodes generating the predictive result. 

In a traditional DNN learning scenario, there exists a training dataset $\mathcal{D} = (x_1, ..., x_n)$ and a loss function $\mathcal{L}$. Each $x_i \in \mathcal{D}$ is defined as a set of features $\mathbf{f}_i$ and a class label $c_i \in \mathcal{C}$ where $\mathcal{C}$ is the set of all possible class values. The final layer of a DNN architecture for such a dataset will consequently contain $|\mathcal{C}|$ nodes, each corresponding to a different class in $\mathcal{C}$. The loss of this DNN given parameters $\theta$ on $\mathcal{D}$ is denoted: $\mathcal{L} = \frac{1}{n}\sum_i^n \mathcal{L}(\theta, x_i)$. 

When $\mathbf{f}_i$ is fed through the DNN with model parameters $\theta$, the output is a set of predicted probabilities $\mathbf{p}_i$. Each value $p_{c,i} \in \mathbf{p}_i$ is the predicted probability that $x_i$ has a class value $c$, and $\mathbf{p}_i$ contains a probability $p_{c,i}$ for each class value $c \in \mathcal{C}$. Each predicted probability $p_{c,i}$ is computed by a node $n_c$ in the final layer of the DNN architecture using input received from the preceding layer and $n_c$'s corresponding parameters in $\theta$. The predicted class for instance $x_i$ given a model $M$ with parameters $\theta$ then becomes $M_\theta(x_i) = \argmax_{c\in\mathcal{C}} p_{c,i}$. Given a cross entropy loss function, the loss on $x_i$ can consequently can be calculated as $\mathcal{L}(\theta, x_i) = -\sum_{c\in\mathcal{C}}y_{c,i}\log(p_{c,i})$ where $y_{c,i} = 1$ if $c=c_i$ and $0$ otherwise. The goal of training a DNN model then becomes to find the parameter values for $\theta$ which minimize the chosen loss function $\mathcal{L}$.

The process of minimizing this loss is typically done through an iterative process called stochastic gradient descent (SGD). At each step, the SGD algorithm (1) selects a batch of samples $B \subseteq \mathcal{D}$, (2) computes the corresponding gradient $\mathbf{g}_B = \frac{1}{|B|}\sum_{x\in B}\nabla_\theta \mathcal{L}(\theta, x)$, and (3) then updates $\theta$ in the direction $-\mathbf{g}_B$. In practice, $\mathcal{D}$ is shuffled and then evenly divided into $|B|$ sized batches such that each sample occurs in exactly one batch. Applying SGD iteratively to each of the pre-determined batches is then referred to as one epoch.

In FL environments however, the training dataset $\mathcal{D}$ is not wholly available at the aggregator. Instead, $N$ participants $\mathcal{P}$ each hold their own private training dataset $D_1, ..., D_N$. Rather than sharing their private raw data, participants instead execute the SGD training algorithm locally and then upload updated parameters to a centralized server (aggregator). Specifically, in the initialization phase (i.e., round 0), the aggregator generates a DNN architecture with parameters $\theta_0$ which is advertised to all participants. At each global training round $r$, a subset $\mathcal{P}_r$ consisting of $k \leq N$ participants is selected based on availability. Each participant $P_i \in \mathcal{P}_r$ executes one epoch of SGD locally on $D_i$ to obtain updated parameters $\theta_{r,i}$, which are sent to the aggregator. The aggregator sets the global parameters $\theta_r = \frac{1}{k}\sum_i \theta_{r, i}$ $\forall i$ where $P_i \in \mathcal{P}_r$. The global parameters $\theta_r$ are then advertised to all $N$ participants. These global parameters at the end of round $r$ are used in the next training round $r+1$. After $R$ total global training rounds, the model $M$ is finalized with parameters $\theta_R$.

\vspace{-8pt}
\subsection{Threat and Adversary Model} \label{subsec:poisoning_fl}
\vspace{-6pt}

\textbf{Threat Model:} We consider the scenario in which a subset of FL participants are malicious or are controlled by a malicious adversary. We denote the percentage of malicious participants among all participants $\mathcal{P}$ as $m\%$. 
Malicious participants may be injected to the system by adding adversary-controlled devices, compromising $m\%$ of the benign participants' devices, or incentivizing (bribing) $m\%$ of benign participants to poison the global model for a certain number of FL rounds. We consider the aggregator to be honest and not compromised.

\textbf{Adversarial Goal:} The goal of the adversary is to manipulate the learned parameters such that the final global model $M$ has high errors for particular classes (a subset of $\mathcal{C}$). The adversary is thereby conducting a targeted poisoning attack. This differs from untargeted attacks which instead seek indiscriminate high global model errors across all classes \cite{biggio2012poisoning,fang2019local,xiao2015feature}. Targeted attacks have the desirable property that they decrease the possibility of the poisoning attack being detected by minimizing influence on non-targeted classes. 

\textbf{Adversary Knowledge and Capability:} We consider a realistic adversary model with the following constraints. Each malicious participant can manipulate the training data $D_i$ on their own device, but 
cannot access or manipulate other participants' data or the model learning process, e.g., SGD implementation, loss function, or server aggregation process. The attack is not specific to the DNN architecture, loss function or optimization function being used. It requires training data to be corrupted, but the learning algorithm remains unaltered.

\vspace{-8pt}
\subsection{Label Flipping Attacks in Federated Learning} \label{subsec:labelflip}
\vspace{-6pt}

We use a label flipping attack to implement targeted data poisoning in FL. Given a source class $c_{src}$ and a target class $c_{target}$ from $\mathcal{C}$, each malicious participant $P_i$ modifies their dataset $D_i$ as follows: For all instances in $D_i$ whose class is $c_{src}$, change their class to $c_{target}$. We denote this attack by $c_{src} \rightarrow c_{target}$. For example, in CIFAR-10 image classification, airplane $\rightarrow$ bird denotes that images whose original class labels are \textit{airplane} will be poisoned by malicious participants by changing their class to \textit{bird}. The goal of the attack is to make the final global model $M$ more likely to misclassify airplane images as bird images at test time.

Label flipping is a well-known attack in centralized ML \cite{shen2016auror,steinhardt2017certified,xiao2012adversarial,xiao2015support}. It is also suitable for the FL scenario given the adversarial goal and capabilities above. Unlike 
other types of poisoning attacks, label flipping does not require the adversary to know the global distribution of $\mathcal{D}$, the DNN architecture, loss function $\mathcal{L}$, etc. It is time and energy-efficient, an attractive feature considering FL is often executed on edge devices
. It is also easy to carry out for non-experts and does not require modification or tampering with participant-side FL software.

\vspace{-16pt}
\begin{table}[!ht]
    \centering
    \begin{tabular}{|x{2cm}|x{\dimexpr\textwidth-3.00cm}|}
        \hline
        $M$, $M_{NP}$ & Model, model trained with no poisoning\\
        \hline
        $k$ & Number of FL participants in each round\\
        \hline
        $R$ & Total number of rounds of FL training\\
        \hline
        $\mathcal{P}_r$ & FL participants queried at round $r$, $r \in [1,R]$\\
        \hline
        $\theta_r, \theta_{r, i}$ & Global model parameters after round $r$ and local model parameters at participant $P_i$ after round $r$\\
        \hline
        $m\%$ & Percentage of malicious participants\\
        \hline
        $c_{src}$, $c_{target}$ & Source and target class in label flipping attack\\
        \hline
        $M^{acc}$ & Global model accuracy\\
        \hline
        $c_i^{recall}$ & Class recall for class $c_i$\\
        \hline
        $m\_cnt^i_j$ & Baseline misclassification count from class $c_i$ to class $c_j$\\
        \hline
    \end{tabular}
    \caption{Notations used throughout the paper.}
    \label{tab:notation}
    \vspace{-28pt}
\end{table}

\textbf{Attack Evaluation Metrics:} At the end of $R$ rounds of FL, the model $M$ is finalized with parameters $\theta_R$. Let $\mathcal{D}_{test}$ denote the test dataset used in evaluating $M$, where $\mathcal{D}_{test} \cap D_i = \emptyset$ for all participant datasets $D_i$. In the next sections, we provide a thorough analysis of label flipping attacks in FL. To do so, we use a number of evaluation metrics.\newline
\textit{Global Model Accuracy} ($M^{acc}$): The global model accuracy is the percentage of instances $x \in \mathcal{D}_{test}$ where the global model $M$ with final parameters $\theta_R$ predicts $M_{\theta_R}(x) = c_i$ and $c_i$ is indeed the true class label of $x$. \newline
\textit{Class Recall} ($c_i^{recall}$): For any class $c_i \in \mathcal{C}$, its class recall is the percentage $\frac{TP_i}{TP_i + FN_i} \cdot 100\%$ where $TP_i$ is the number of instances $x \in \mathcal{D}_{test}$ where $M_{\theta_R}(x) = c_i$ \textbf{and} $c_i$ is the true class label of $x$; and $FN_i$ is the number of instances $x \in \mathcal{D}_{test}$ where $M_{\theta_R}(x) \neq c_i$ and the true class label of $x$ is $c_i$. \newline
\textit{Baseline Misclassification Count} ($m\_cnt^i_j$): Let $M_{NP}$ be a global model trained for $R$ rounds using FL without any malicious attack. For 
classes $c_i \neq c_j$, the baseline misclassification count from 
$c_i$ to 
$c_j$, denoted $m\_cnt^i_j$, is defined as the number of instances $x \in \mathcal{D}_{test}$ where $M_{NP}(x) = c_j$ \textbf{and} the true class 
of $x$ is $c_i$.

Table~\ref{tab:notation} provides a summary of the notation used in the rest of this paper.

\vspace{-12pt}
\section{Analysis of Label Flipping Attacks in FL} \label{sec:experiments}
\vspace{-6pt}
\subsection{Experimental Setup}\label{subsec:exp_setup}
\vspace{-6pt}

\textbf{Datasets and DNN Architectures:} We conduct our attacks using two popular image classification datasets: CIFAR-10 \cite{krizhevsky2009learning} and Fashion-MNIST \cite{xiao2017fashion}. CIFAR-10 consists of 60,000 color images in 10 object classes such as deer, airplane, and dog with 6,000 images included per class. The complete dataset is pre-divided into 50,000 training images and 10,000 test images. Fashion-MNIST consists of a training set of 60,000 images and a test set of 10,000 images. Each image in Fashion-MNIST is gray-scale and associated with one of 10 classes of clothing such as pullover, ankle boot, or bag. In experiments with CIFAR-10, we use a convolutional neural network with six convolutional layers, batch normalization, and two fully connected dense layers. This DNN architecture achieves a test accuracy of 79.90\% in the centralized learning scenario, i.e. $N=1$, without poisoning. In experiments with Fashion-MNIST, we use a two layer convolutional neural network with batch normalization, an architecture which achieves 91.75\% test accuracy in the centralized scenario without poisoning. Further details of the datasets and DNN model architectures can be found in Appendix \ref{appendix:a}.

\textbf{Federated Learning Setup:} We implement FL in Python using the PyTorch~\cite{paszke2019pytorch} library. By default, we have $N=50$ participants, one central aggregator, and $k=5$. We use an \textit{independent and identically distributed} (\textit{iid}) data distribution, i.e., we assume the total training dataset is uniformly randomly distributed among all participants with each participant receiving a unique subset of the training data. The testing data is used for model evaluation only and is therefore not included in any participant $P_i$'s train dataset $D_i$. Observing that both DNN models converge after fewer than 200 training rounds, we set our FL experiments to run for $R=200$ rounds total.

\textbf{Label Flipping Process:} In order to simulate the label flipping attack in a FL system with $N$ participants of which $m\%$ are malicious, at the start of each experiment we randomly designate $N \times m\%$ of the participants from $\mathcal{P}$ as malicious. The rest are honest. To address the impact of random selection of malicious participants, by default we repeat each experiment 10 times and report the average results. Unless otherwise stated, we use $m = 10\%$. 

For both datasets we consider three label flipping attack settings representing a diverse set of conditions in which to base adversarial attacks. These conditions include (1) a source class $\rightarrow$ target class pairing whose source class was very frequently misclassified as the target class in federated, non-poisoned training, (2) a pairing where the source class was very infrequently misclassified as the target class, and (3) a pairing between these two extremes. Specifically, for CIFAR-10 we test (1) 5: dog $\rightarrow$ 3: cat, (2) 0: airplane $\rightarrow$ 2: bird, and (3) 1: automobile $\rightarrow$ 9: truck. For Fashion-MNIST we experiment with (1) 6: shirt $\rightarrow$ 0: t-shirt/top, (2) 1: trouser $\rightarrow$ 3: dress, and (3) 4: coat $\rightarrow$ 6: shirt.

\vspace{-8pt}
\subsection{Label Flipping Attack Feasibility} \label{subsec:feasibility}
\vspace{-6pt}

\begin{figure}[!t]
  \begin{subfigure}{0.24\textwidth}
    \centering\includegraphics[width=\textwidth]{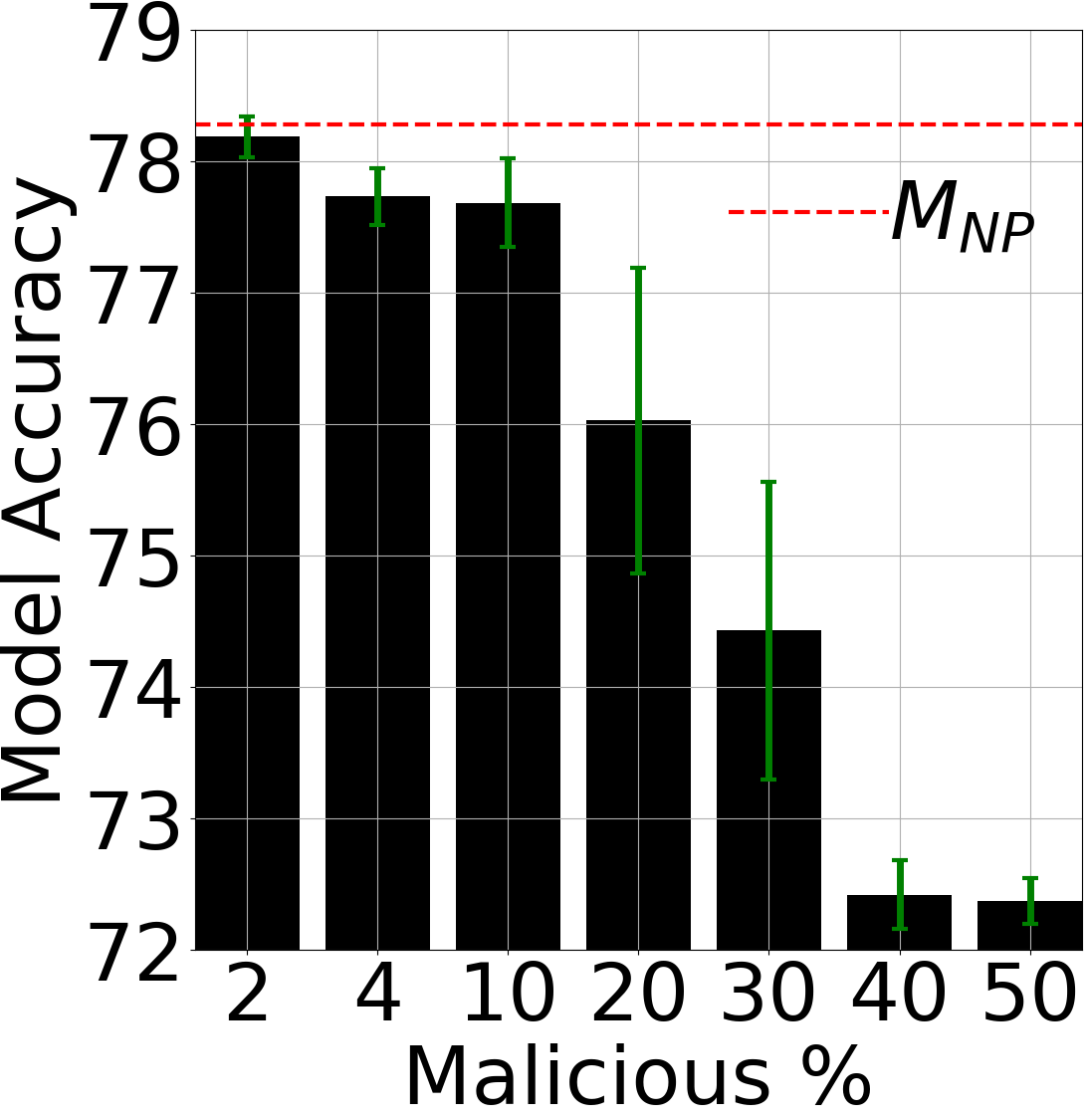}
    \caption{\scriptsize{CIFAR-10 $M^{acc}$}}
    \label{fig:1Aa}
  \end{subfigure}
  \begin{subfigure}{0.24\textwidth}
    \centering\includegraphics[width=\textwidth]{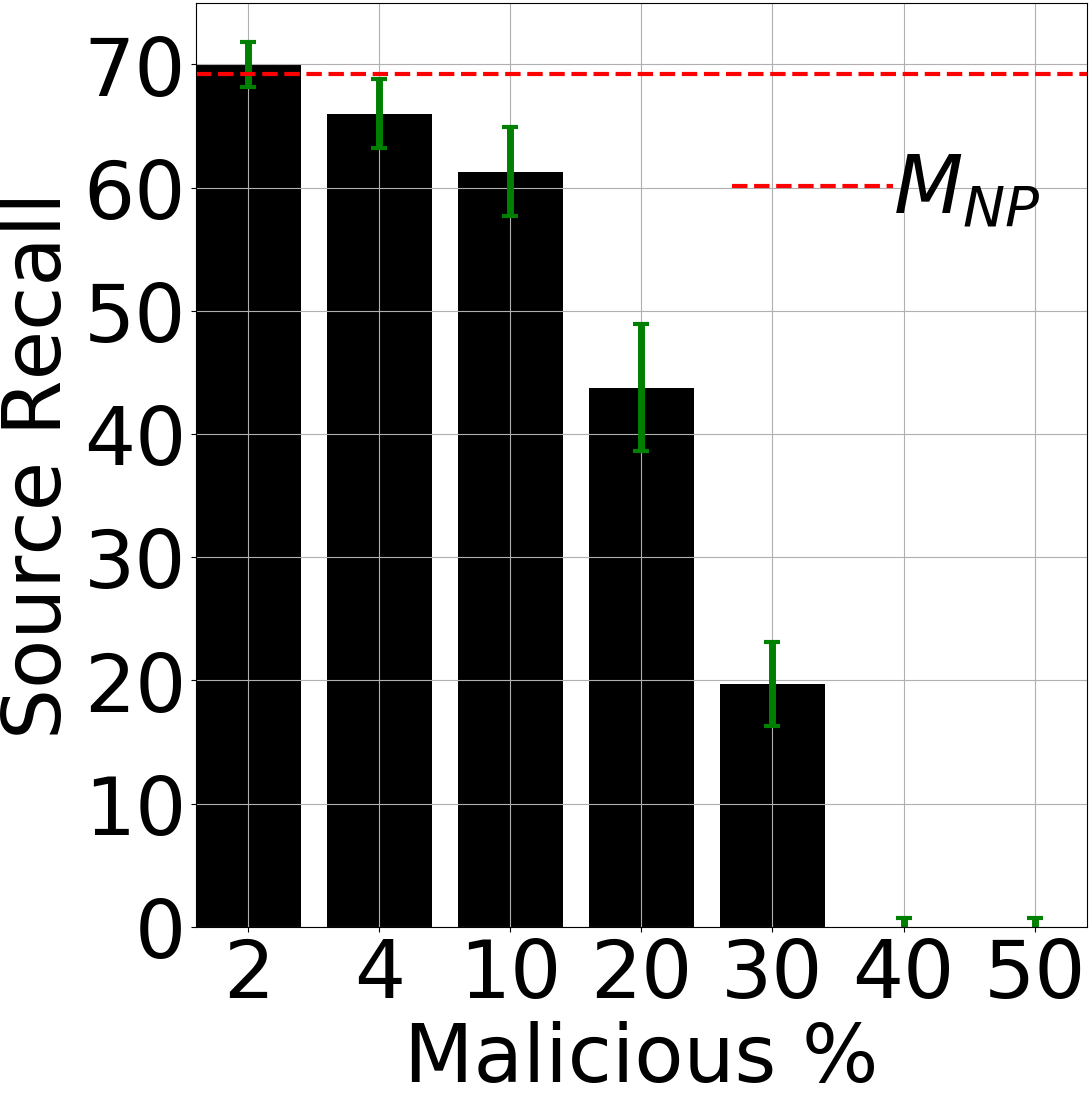}
    \caption{\scriptsize{CIFAR-10 $c_{src}^{recall}$}}
    \label{fig:1Ab}
  \end{subfigure}
  \begin{subfigure}{0.24\textwidth}
    \centering\includegraphics[width=\textwidth]{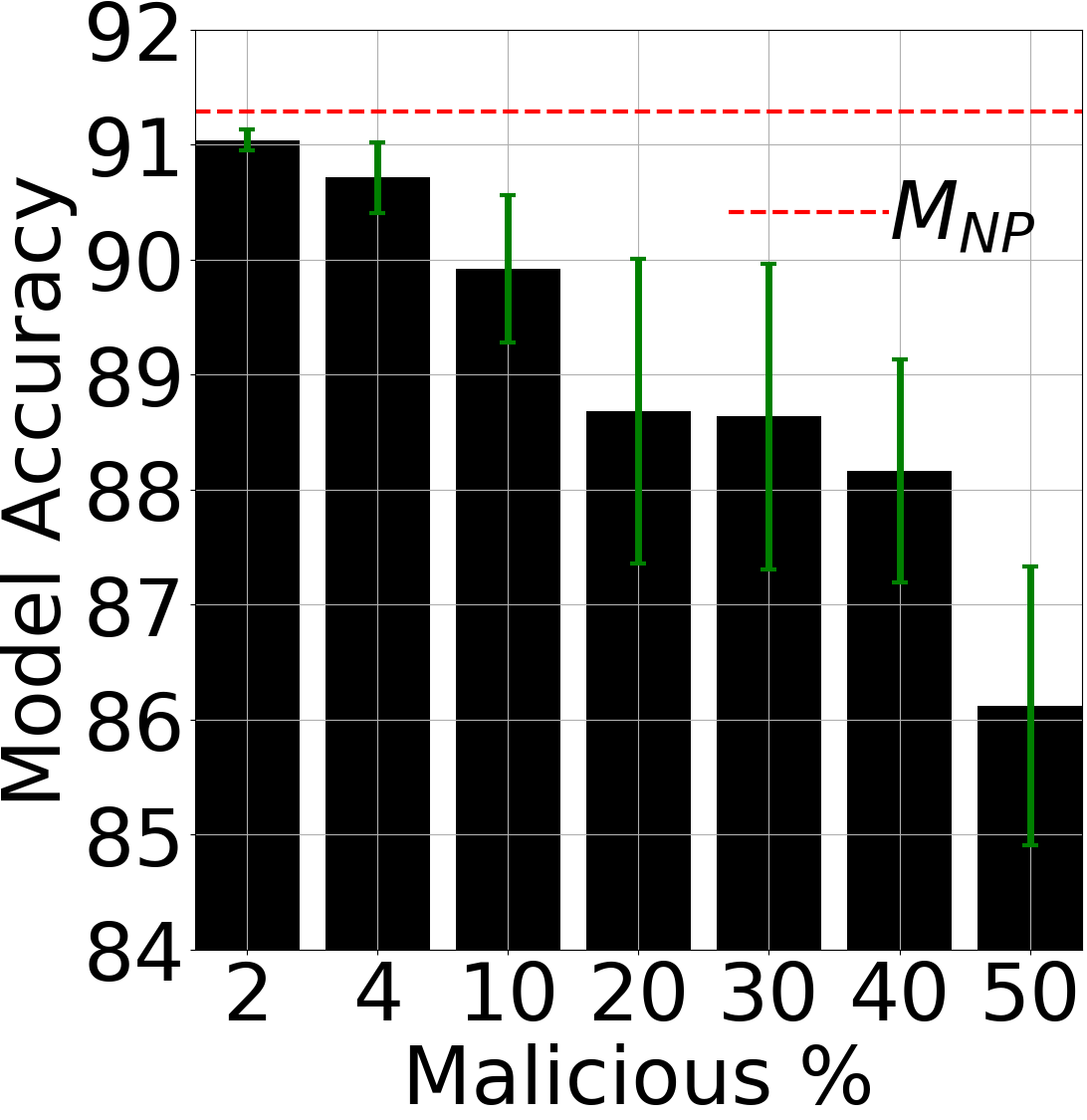}
    \caption{\scriptsize{F-MNIST $M^{acc}$}}
    \label{fig:1Ac}
  \end{subfigure}
  \begin{subfigure}{0.24\textwidth}
    \centering\includegraphics[width=\textwidth]{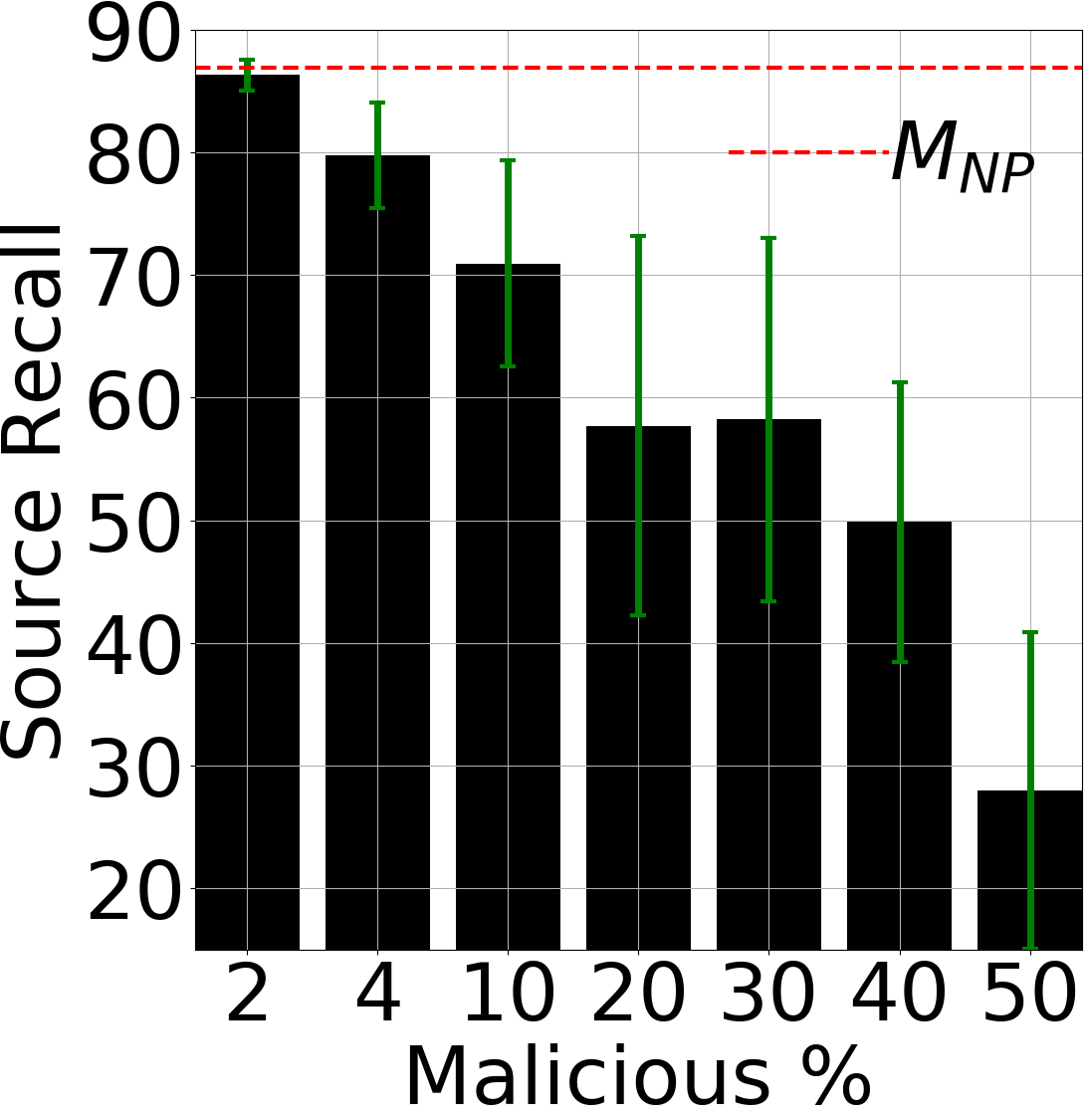}
    \caption{\scriptsize{F-MNIST $c_{src}^{recall}$}}
    \label{fig:1Ad}
  \end{subfigure}
  \caption{Evaluation of attack feasibility and impact of malicious participant percentage on attack effectiveness. CIFAR-10 experiments are for the 5 $\rightarrow$ 3 setting while Fashion-MNIST experiments are for the 4 $\rightarrow$ 6 setting. Results are averaged from 10 runs for each setting of $m\%$. The black bars are mean over the 10 runs and the green error bars denote standard deviation.}
  \label{fig:1A}
  \vspace{-10pt}
\end{figure}

We start by investigating the feasibility of poisoning FL systems using label flipping attacks. Figure~\ref{fig:1A} outlines the global model accuracy and source class recall in scenarios with malicious participant percentage $m$ ranging from 2\% to 50\%. Results demonstrate that as the malicious participant percentage, increases the global model utility (test accuracy) decreases. Even with small $m$, we observe a decrease in model accuracy compared to a non-poisoned model (denoted by $M_{NP}$ in the graphs), and there is an even larger decrease in source class recall. In experiments with CIFAR-10, once $m$ reaches 40\%, the recall of the source class decreases to 0\% and the global model accuracy decreases from 78.3\% in the non-poisoned setting to 74.4\% in the poisoned setting. Experiments conducted on Fashion-MNIST show a similar pattern of utility loss. With $m=4\%$ source class recall drops by $\sim 10\%$ and with $m=10\%$ it drops by $\sim 20\%$. It is therefore clear that an adversary who controls even a minor proportion of the total participant population is capable of significantly impacting global model utility.

While both datasets are vulnerable to label flipping attacks, the degree of vulnerability varies between datasets with CIFAR-10 demonstrating more vulnerability than Fashion-MNIST. For example, consider the 30\% malicious scenario, Figure~\ref{fig:1Ab} shows the source class recall for the CIFAR-10 dataset drops to 19.7\% while Figure~\ref{fig:1Ad} shows a much lower decrease for the Fashion-MNIST dataset with 58.2\% source class recall under the same experimental settings. 

\begin{table}[!t]
    \centering
    \resizebox{\textwidth}{!}{
    \begin{tabular}{|ccc|c|c|c|c|c|c|c|c|}
        \hline
        \multirow{2}{*}{\shortstack{$c_{src}$}} & \multirow{2}{*}{$\rightarrow$} & \multirow{2}{*}{\shortstack{$c_{target}$}} & \multirow{2}{*}{\shortstack{$m\_cnt^{src}_{target}$}} & \multicolumn{7}{c|}{Percentage of Malicious Participants ($m\%$)} \\
        \cline{5-11}
        & & & & 2 & 4 & 10 & 20 & 30 & 40 & 50 \\
        \hline
        \multicolumn{11}{|c|}{CIFAR-10} \\
        \hline
        0 & $\rightarrow$ & 2 & 16 & \cellcolor{yellow}{\textbf{1.42\%}} & 2.93\% & \cellcolor{yellow}{\textbf{10.2\%}} & 14.1\% & 48.3\% & \cellcolor{yellow}{\textbf{73\%}} & \cellcolor{yellow}{\textbf{70.5\%}} \\
        1 & $\rightarrow$ & 9 & 56 & 0.69\% & \cellcolor{yellow}{\textbf{3.75\%}} & 6.04\% & 15\% & 36.3\% & 49.2\% & 54.7\% \\
        5 & $\rightarrow$ & 3 & 200 & 0\% & 3.21\% & 7.92\% & \cellcolor{yellow}{\textbf{25.4\%}} & \cellcolor{yellow}{\textbf{49.5\%}} & 69.2\% & 69.2\% \\
        \hline
        \multicolumn{11}{|c|}{Fashion-MNIST} \\
        \hline
        1 & $\rightarrow$ & 3 & 18 & 0.12\% & 0.42\% & 2.27\% & 2.41\% & \cellcolor{yellow}{\textbf{40.3\%}} & \cellcolor{yellow}{\textbf{45.4\%}} & 42\% \\
        4 & $\rightarrow$ & 6 & 51 & \cellcolor{yellow}{\textbf{0.61\%}} & \cellcolor{yellow}{\textbf{7.16\%}} & \cellcolor{yellow}{\textbf{16\%}} & \cellcolor{yellow}{\textbf{29.2\%}} & 28.7\% & 37.1\% & \cellcolor{yellow}{\textbf{58.9\%}} \\
        6 & $\rightarrow$ & 0 & 118 & -1\% & 2.19\% & 7.34\% & 9.81\% & 19.9\% & 39\% & 43.4\% \\
        \hline
    \end{tabular}
    }
    \caption{Loss in source class recall for three source $\rightarrow$ target class settings with differing baseline misclassification counts in CIFAR-10 and Fashion-MNIST. Loss averaged from 10 runs. Highlighted bold entries 
    are 
    highest loss in 
    each.}
    \label{tab:1B}
    \vspace{-30pt}
\end{table}

On the other hand, vulnerability variation based on source and target class settings is less clear. In Table \ref{tab:1B}, we report the results of three different combinations of source $\rightarrow$ target attacks for each dataset. Consider the two extreme settings for the CIFAR-10 dataset: on the low end the 0 $\rightarrow$ 2 setting has a baseline misclassification count of 16 while the high end count is 200 for the 5 $\rightarrow$ 3 setting. Because of the DNN's relative challenge in differentiating class 5 from class 3 in the non-poisoned setting, it could be anticipated that conducting a label flipping attack within the 5 $\rightarrow$ 3 setting would result in the greatest impact on source class recall. However, this was not the case. Table~\ref{tab:1B} shows that in only two out of the six experimental scenarios did 5 $\rightarrow$ 3 record the largest drop in source class recall. In fact, four scenarios' results show the 0 $\rightarrow$ 2 setting, the setting with the lowest baseline misclassification count, as the most effective option for the adversary. Experiments with Fashion-MNIST show a similar trend, with label flipping attacks conducted in the 4 $\rightarrow$ 6 setting being the most successful rather than the 6 $\rightarrow$ 0 setting which has more than $2\times$ the number of baseline misclassifications. These results indicate that identifying the most vulnerable source and target class combination may be a non-trivial task for the adversary, and that there is not necessarily a correlation between non-poisoned misclassification performance and attack effectiveness.

\begin{table}[!t]
    \centering
    \begin{tabular}{|ccc|x{0.15\textwidth}|x{0.15\textwidth}|c|c|c|}
        \hline 
        \multirow{2}{*}{\shortstack{$c_{src}$}} & \multirow{2}{*}{$\rightarrow$} & \multirow{2}{*}{\shortstack{$c_{target}$}} & \multirow{2}{*}{$\Delta ~ c_{src}^{recall}$} & 
        \multirow{2}{*}{$\Delta ~ c_{target}^{recall}$} & \multirow{2}{*}{\shortstack{$\sum$ all other $\Delta ~ c^{recall}$}} \\ 
        & & & & & \\ 
        \hline
        \multicolumn{6}{|c|}{CIFAR-10} \\
        \hline
        0 & $\rightarrow$ & 2 & -6.28\% & 1.58\% & 0.34\% \\
        1 & $\rightarrow$ & 9 & -6.22\% & 2.28\% & 0.16\% \\
        5 & $\rightarrow$ & 3 & -6.12\% & 3.00\% & 0.17\% \\
        \hline
        \multicolumn{6}{|c|}{Fashion-MNIST} \\
        \hline
        1 & $\rightarrow$ & 3 & -2.23\% & 0.25\% & 0.01\% \\
        4 & $\rightarrow$ & 6 & -9.96\% & 2.40\% & 0.09\% \\
        6 & $\rightarrow$ & 0 & -8.87\% & 2.59\% & 0.20\% \\
        \hline
    \end{tabular}
    \caption{Changes due to poisoning in source class recall, target class recall, and total recall for all remaining classes (non-source, non-target). Results are averaged from 10 runs in each setting. The maximum standard deviation observed was 1.45\% in source class recall and 1.13\% in target class recall.} 
    \label{tab:1C}
    \vspace{-28pt}
\end{table}

\begin{figure}[h]
\vspace{-12pt}
\centering
  \begin{subfigure}{0.44\textwidth}
    \centering\includegraphics[width=\textwidth]{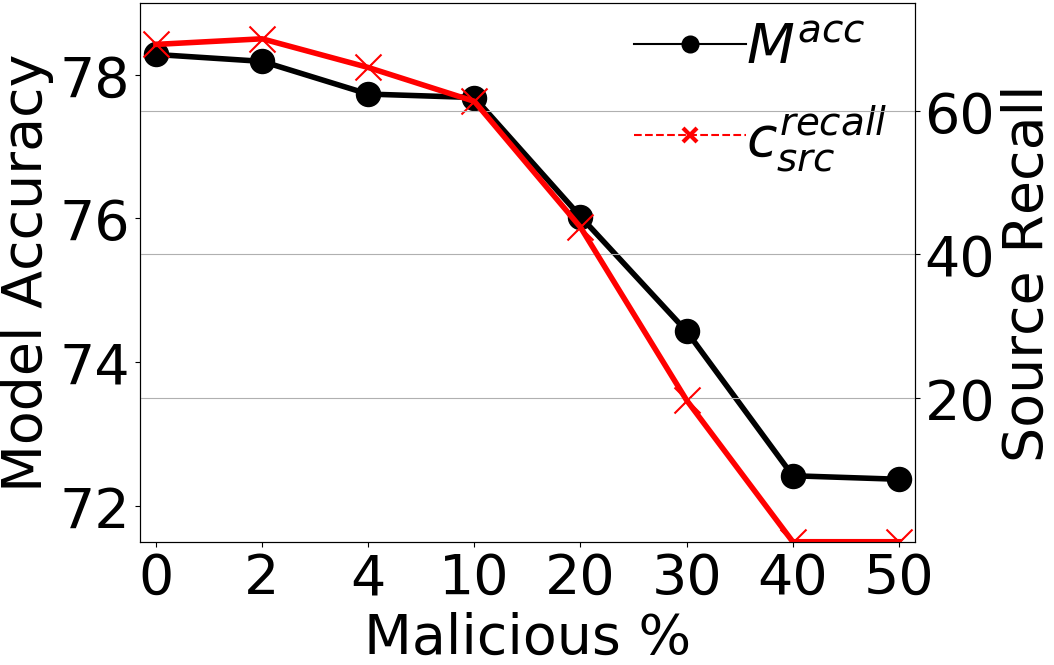}
    \caption{CIFAR-10}
  \end{subfigure}
  \hspace{6mm}
  \begin{subfigure}{0.44\textwidth}
    \centering\includegraphics[width=\textwidth]{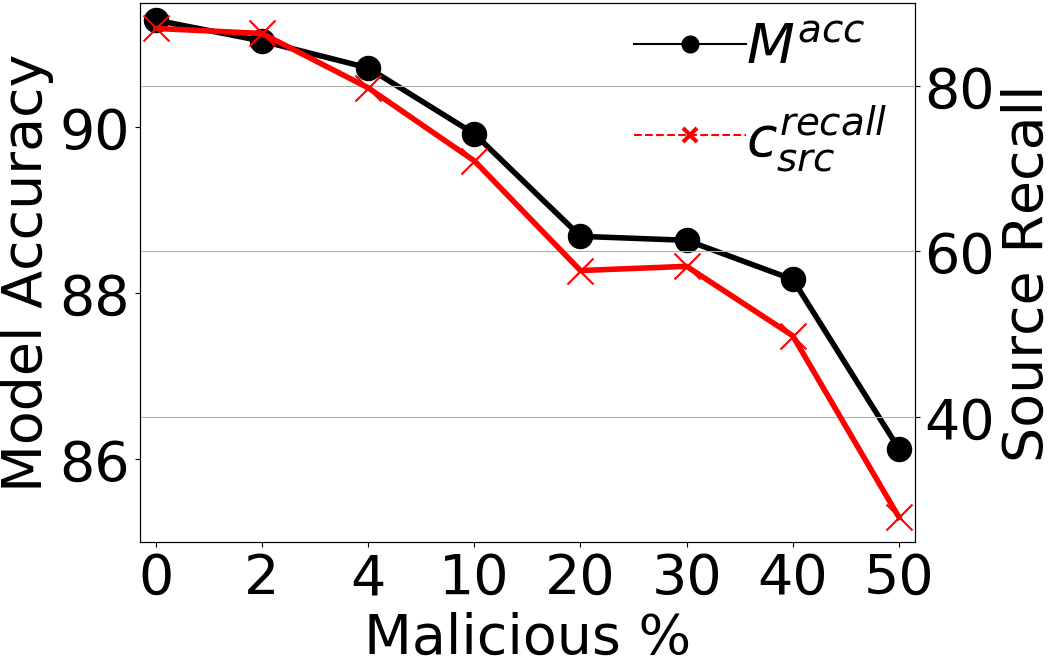}
    \caption{Fashion-MNIST}
  \end{subfigure}
  \caption{Relationship between global model accuracy and source class recall across changing percentages of malicious participants for CIFAR-10 and Fashion-MNIST. As each dataset has 10 classes, the scale for $M^{acc}$ vs $c_{src}^{recall}$ is 1:10.} 
  \label{fig:1C}
\vspace{-18pt}
\end{figure}

We additionally study a desirable feature of the label flipping attack: they appear to be targeted. Specifically, Table~\ref{tab:1C} reports the following quantities for each source $\rightarrow$ target flipping scenario: loss in source class recall, loss in target class recall, and loss in recall of all remaining classes. We observe that the attack causes substantial change in source class recall ($> 6\%$ drop in most cases) and target class recall. However, the attack impact on the recall of remaining classes is an order of magnitude smaller. CIFAR-10 experiments show a maximum of 0.34\% change in class recalls attributable to non-source and non-target classes and Fashion-MNIST experiments similarly show a maximum change of 0.2\% attributable to non-source and non-target classes, both of which are relatively minor compared to source and target classes. Thus, the attack is causing the global model to misclassify instances belonging to $c_{src}$ as $c_{target}$ at test time while other classes remain relatively unimpacted, demonstrating its targeted nature towards $c_{src}$ and $c_{target}$. Considering the large impact of the attack on source class recall, changes in source class recall therefore make up the vast majority of the decreases in global model accuracy caused by label flipping attacks in FL systems. This observation can also be seen in Figure~\ref{fig:1C} where the change in global model accuracy closely follows the change in source class recall.

The targeted nature of the label flipping attack allows for adversaries to remain under the radar in many FL systems. Consider systems where the data contain 100 classes or more, as is the case in CIFAR-100~\cite{krizhevsky2009learning} and ImageNet~\cite{deng2009imagenet}. In such cases, targeted attacks become much more stealthy due to their limited impact to classes other than source and target.

\vspace{-8pt}
\subsection{Attack Timing in Label Flipping Attacks} \label{subsec:timing}
\vspace{-6pt}

While label flipping attacks can occur at any point in the learning process and last for arbitrary lengths, it is important to understand the capabilities of adversaries who are available for only part of the training process. For instance, Google's Gboard application of FL requires all participant devices be plugged into power and connected to the internet via WiFi \cite{47976}. Such requirements create cyclic conditions where many participants are not available during the day, when phones are not plugged in and are actively in use. Adversaries can take advantage of this design choice, making themselves available at times when honest participants are unable to.

\begin{figure}[!t]
\centering
  \begin{subfigure}{0.44\textwidth} 
    \centering\includegraphics[width=\textwidth]{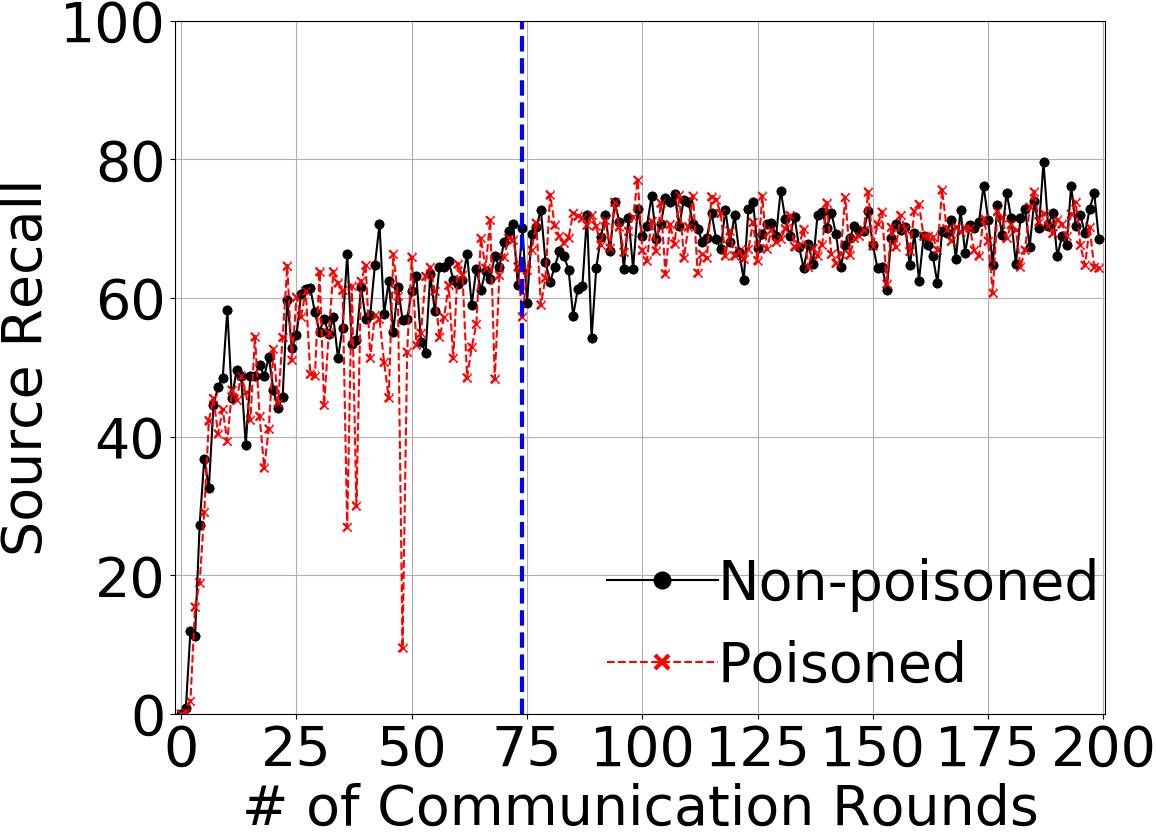}
    \caption{CIFAR-10}
  \end{subfigure}
    \hspace{6mm}
  \begin{subfigure}{0.44\textwidth}  
    \centering\includegraphics[width=\textwidth]{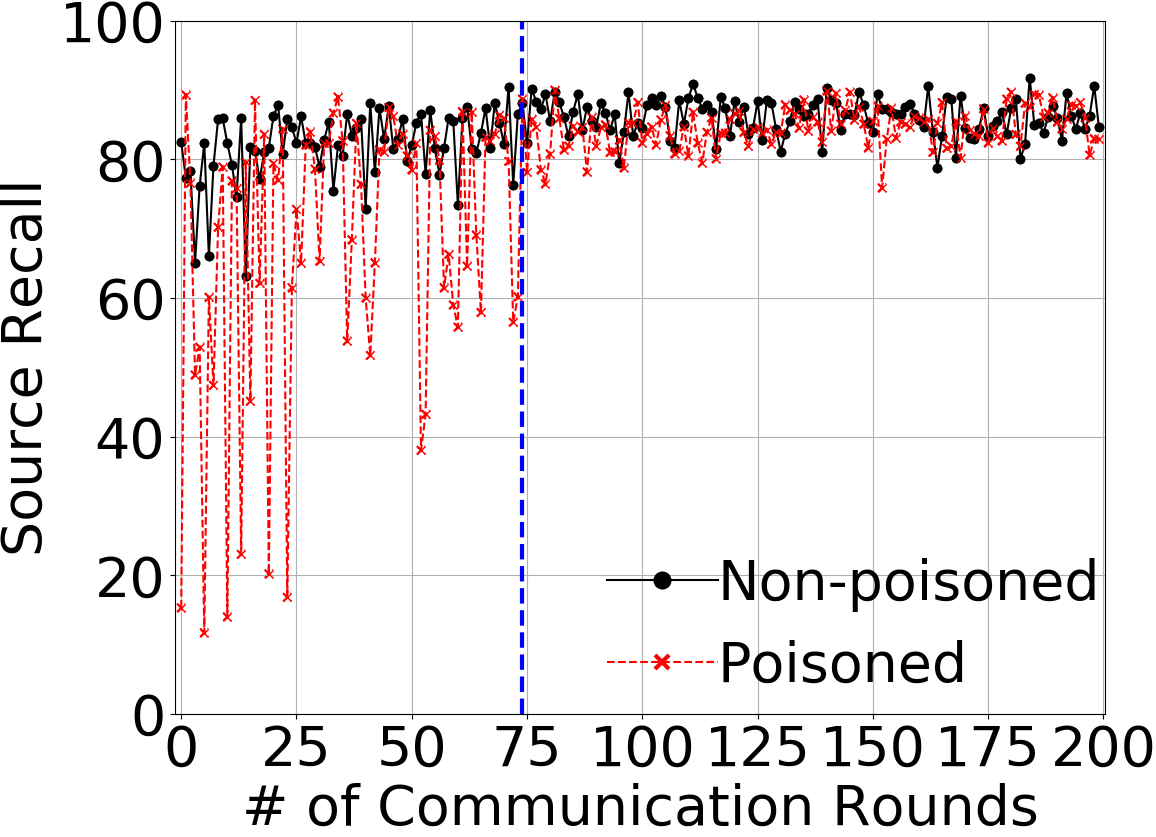}
    \caption{Fashion-MNIST}
  \end{subfigure}
  \caption{Source class recall by round for experiments with ``early round poisoning", i.e., malicious participation only in the first 75 rounds ($r<75$). The blue line indicates the round at which malicious participation is no longer allowed.}
  \label{fig:2A}
  \vspace{-12pt}
\end{figure}

\begin{figure}[!t]
\centering
  \begin{subfigure}{0.44\textwidth}
    \centering\includegraphics[width=\textwidth]{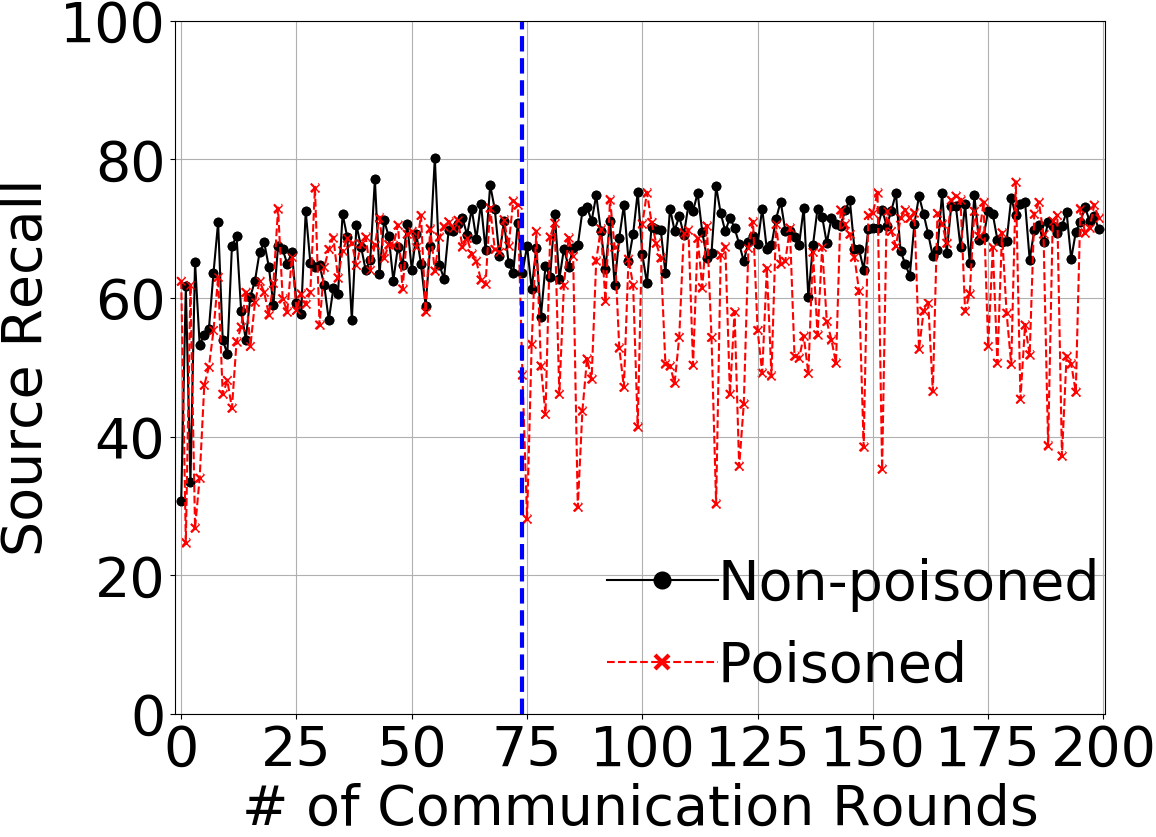}
    \caption{CIFAR-10}
  \end{subfigure}
  \hspace{6mm}
  \begin{subfigure}{0.44\textwidth}
    \centering\includegraphics[width=\textwidth]{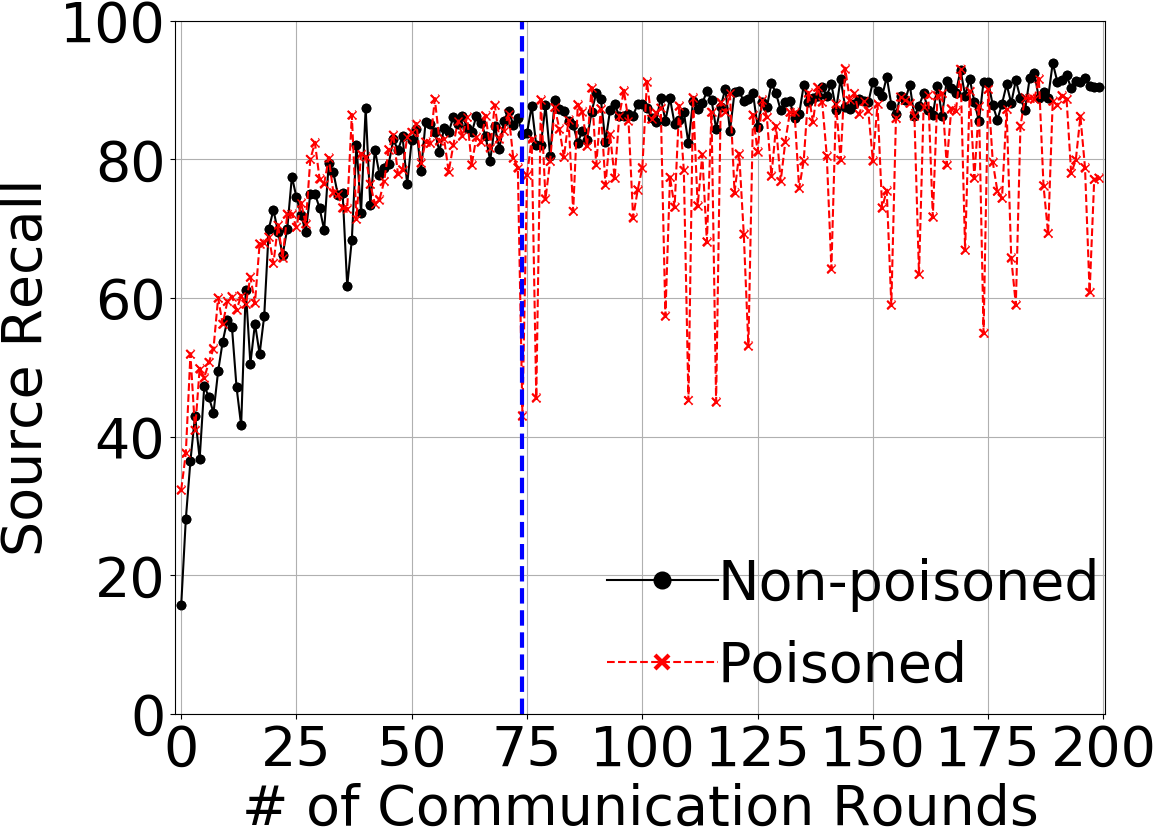}
    \caption{Fashion-MNIST}
  \end{subfigure}
  \caption{Source class recall by round for experiments with ``late round poisoning", i.e., malicious participation only after round 75 ($r\geq75$). The blue line indicates the round at which malicious participation starts.}
  \label{fig:2B}
  \vspace{-18pt}
\end{figure}

We consider two scenarios in which the adversary is restricted in the time in which they are able to make malicious participants available: one in which the adversary makes malicious participants available only before the 75th training round, and one in which malicious participants are available only after the 75th training round. As the rate of global model accuracy improvement decreases with both datasets by training round 75, we choose this point to highlight how pre-established model stability may effect an adversary's ability to launch an effective label flipping attack. Results for the first scenario are given in Figure \ref{fig:2A} whereas the results for the second scenario are given in Figure \ref{fig:2B}. 

In Figure \ref{fig:2A}, we compare source class recall in a non-poisoned setting versus 
with poisoning only before round 75. Results on both CIFAR-10 and Fashion-MNIST show that while there are observable drops in source class recall during the rounds with poisoning (1-75), the global model is able to recover quickly after poisoning finishes (after round 75). Furthermore, the final convergence of the models (towards the end of training
) are not impacted, given the models with and without poisoning are converge with roughly the same recall values. We do note that some CIFAR-10 experiments exhibited delayed convergence by an additional 50-100 training rounds, but these circumstances were rare and still eventually achieved the accuracy and recall levels of a non-poisoned model despite delayed convergence.

\begin{wraptable}[15]{r}{0.55\textwidth}
\vspace{-42pt}
    \begin{center}
    \begin{tabular}{|ccc|c|c|}
        \hline 
        \multirow{2}{*}{\shortstack{$c_{src}$}} & \multirow{2}{*}{$\rightarrow$} & \multirow{2}{*}{\shortstack{$c_{target}$}} & \multicolumn{2}{c|}{Source Class Recall $(c_{src}^{recall})$} \\ 
        \cline{4-5}
        & & & $m\% \in \mathcal{P}_R > 0$ & $m\% \in \mathcal{P}_R = 0$ \\
        \hline
        \multicolumn{5}{|c|}{CIFAR-10} \\
        \hline
        0 & $\rightarrow$ & 2 & 73.90\% & 82.45\% \\
        1 & $\rightarrow$ & 9 & 77.30\% & 89.40\% \\
        5 & $\rightarrow$ & 3 & 57.50\% & 73.10\% \\
        \hline
        \multicolumn{5}{|c|}{Fashion-MNIST} \\
        \hline
        1 & $\rightarrow$ & 3 & 84.32\% & 96.25\% \\
        4 & $\rightarrow$ & 6 & 51.50\% & 89.60\% \\
        6 & $\rightarrow$ & 0 & 49.80\% & 73.15\% \\
        \hline
    \end{tabular}
    \caption{Final source class recall when at least one malicious party participates in the final round $R$ versus when all participants in round $R$ are non-malicious. Results averaged for 10 runs for each experimental setting.}
    \label{tab:2B}
    \end{center}
\end{wraptable}
In Figure \ref{fig:2B}, we compare source class recall in a non-poisoned setting versus with poisoning limited to the 75th and later training rounds. These results show the impact of such late poisoning demonstrating limited longevity; a phenomena which can be seen in the quick and dramatic changes in source class recall. Specifically, source class recall quickly returns to baseline levels once fewer malicious participants are selected in a training round even immediately following a round with a large number of malicious participants having caused a dramatic drop. However, the final poisoned model in the late-round poisoning scenario may show substantial difference in accuracy or recall compared to a non-poisoned model. This is evidenced by the CIFAR-10 experiment in Figure \ref{fig:2B}, in which the source recall of the poisoned model is $\sim$ 10\% lower compared to 
non-poisoned
. 

Furthermore, we observe that model convergence on both data\-sets is negatively impacted, as evidenced by the large variances in recall values between consecutive rounds. Consider Table~\ref{tab:2B} where results are compared when either (1) at least one malicious participant is selected for $\mathcal{P}_R$ or (2) $\mathcal{P}_R$ is made entirely of honest participants. When at least one malicious participant is selected, the final source class recall is, on average, 12.08\% lower with the CIFAR-10 dataset and 24.46\% lower with the Fashion-MNIST dataset. The utility impact from the label flipping attack is therefore predominantly tied to the number of malicious participants selected in the last few rounds of training. 

\vspace{-8pt}
\subsection{Malicious Participant Availability} \label{sec:part-availability}
\vspace{-6pt}

Given the impact of malicious participation in late training rounds on attack effectiveness, we now introduce a malicious participant availability parameter $\alpha$. By varying $\alpha$ we can simulate the adversary's ability to control compromised participants' availability (i.e. ensuring connectivity or power access) at various points in training. Specifically, $\alpha$ represents malicious participants' availability and therefore likeliness to be selected relative to honest participants. For example, if $\alpha=0.6$, when selecting each participant $P_i \in \mathcal{P}_r$ for round $r$, there is a 0.6 probability that $P_i$ will be one of the malicious participants. Larger $\alpha$ implies higher likeliness of malicious participation. In cases where $k > N \times m\%$, the number of malicious participants in $\mathcal{P}_r$ is bounded by $N \times m\%$.

\begin{figure}[!t]
\centering
  \begin{subfigure}{0.47\textwidth}
    \centering\includegraphics[width=0.9\textwidth]{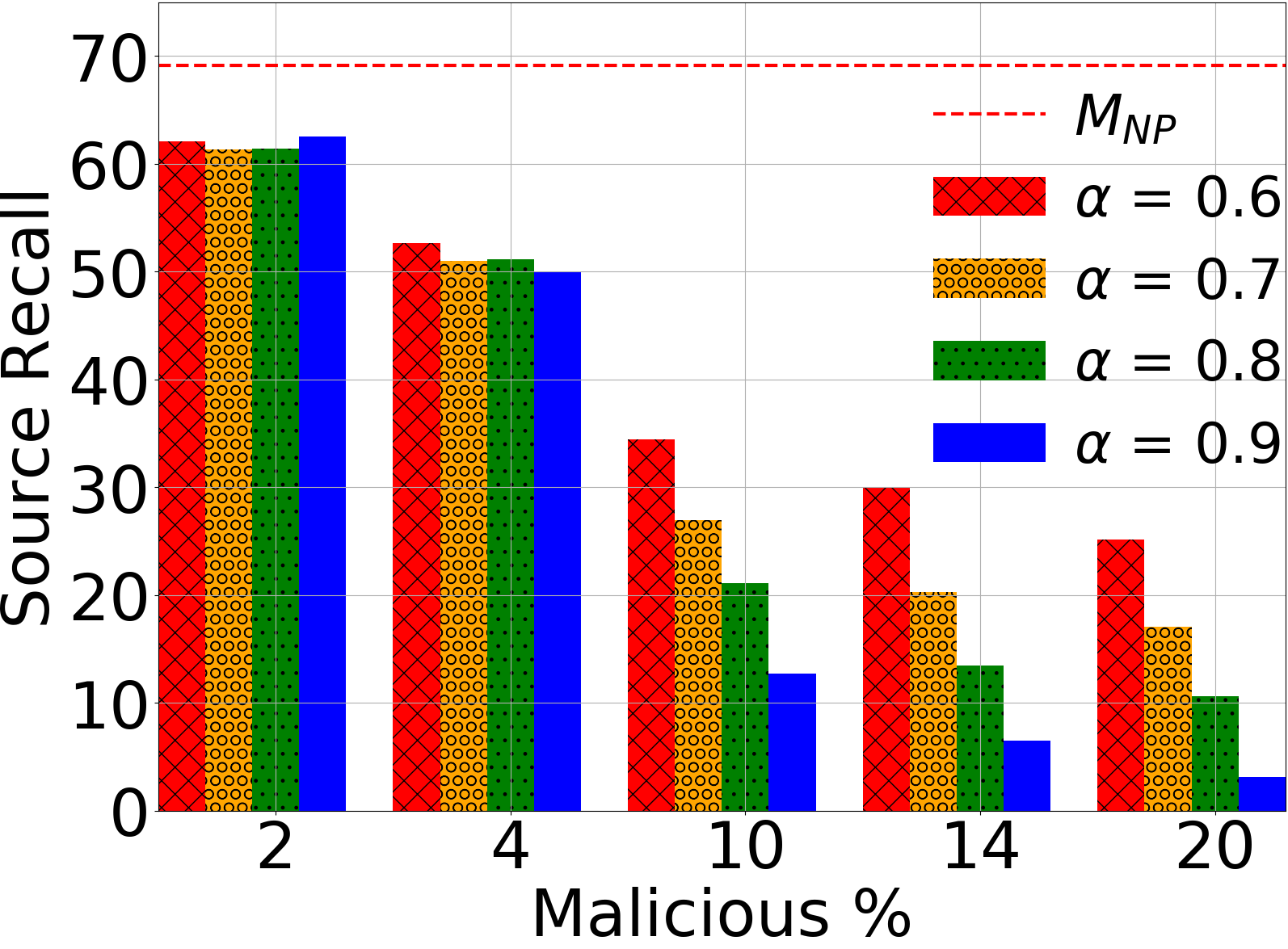}
    \caption{CIFAR-10}
  \end{subfigure}
  \hspace{4mm}
  \begin{subfigure}{0.47\textwidth}
    \centering\includegraphics[width=0.9\textwidth]{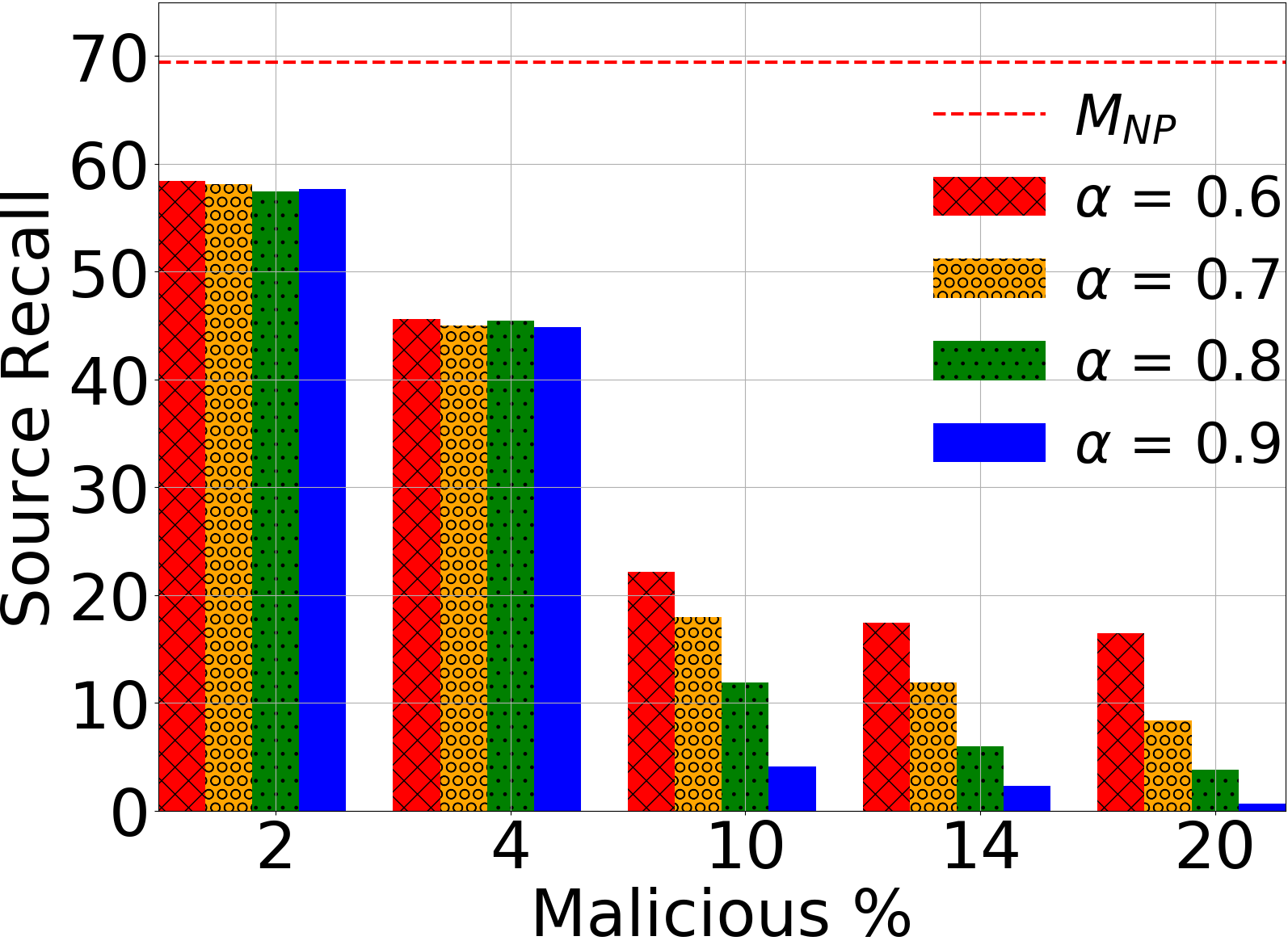}
    \caption{Fashion-MNIST}
  \end{subfigure}
  \caption{Evaluation of impact from malicious participants' availability $\alpha$ on source class recall. Results are averaged from 3 runs for each setting.} 
  \label{fig:3A}
  \vspace{-18pt}
\end{figure}

Figure~\ref{fig:3A} reports results for varying values of $\alpha$ in late round poisoning, i.e., malicious participation is limited to rounds $r \geq 75$. Specifically, we are interested in studying those scenarios where an adversary boosts the availability of the malicious participants enough that their selection becomes more likely than the non-malicious participants, hence in Figure 5 we use $\alpha \geq 0.6$. The reported source class recalls in Figure \ref{fig:3A} are averaged over the last 125 rounds (total 200 rounds minus first 75 rounds) to remove the impact of individual round variability; further, each experiment setting is repeated 3 times and results are averaged. The results show that, when the adversary maintains sufficient representation in the participant pool (i.e. $m \geq 10\%$), manipulating the availability of malicious participants can yield significantly higher impact on the global model utility with source class recall losses in excess of 20\%. On both datasets with $m \geq 10\%$, the negative impact on source class recall is highest with $\alpha=0.9$, which is followed by $\alpha=0.8$, $\alpha=0.7$ and $\alpha=0.6$, i.e., in decreasing order of malicious participant availability. Thus, in order to mount an impactful attack, it is in the best interests of the adversary to perform the attack \textit{with highest malicious participant availability in late rounds}. We note that when $k$ is significantly larger than $N \times m\%$, increasing availability ($\alpha$) will be insufficient for meaningfully increasing malicious participant selection in individual training rounds. Therefore, experiments where $m < 10\%$ show little variation despite changes in $\alpha$.

\begin{figure}[!t]
\centering
  \begin{subfigure}{0.45\textwidth}
    \centering\includegraphics[width=0.95\textwidth]{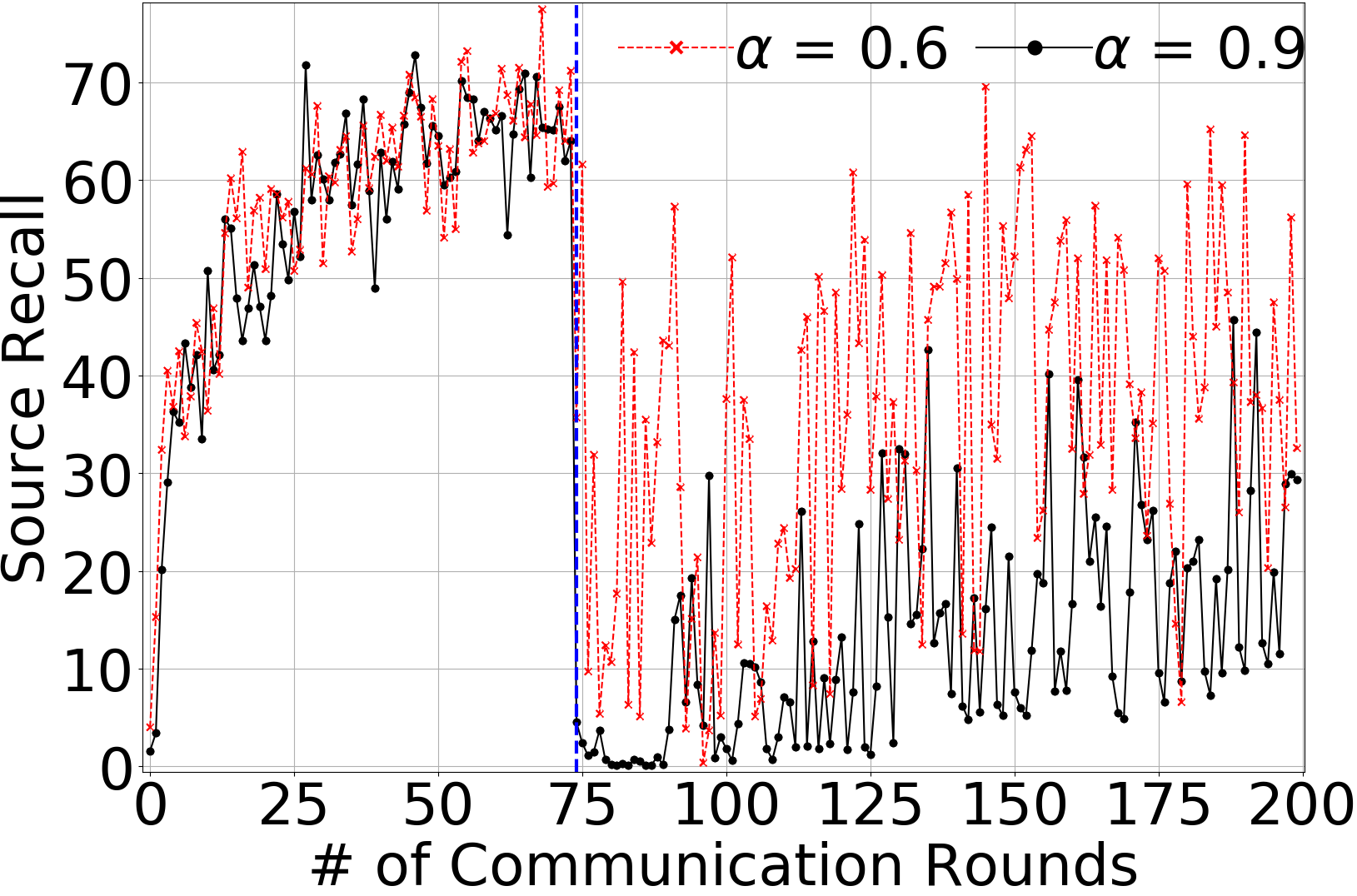}
    \vspace{-3pt}
    \caption{CIFAR-10}
  \end{subfigure}
  \hspace{6pt}
  \begin{subfigure}{0.45\textwidth}
    \centering\includegraphics[width=0.95\textwidth]{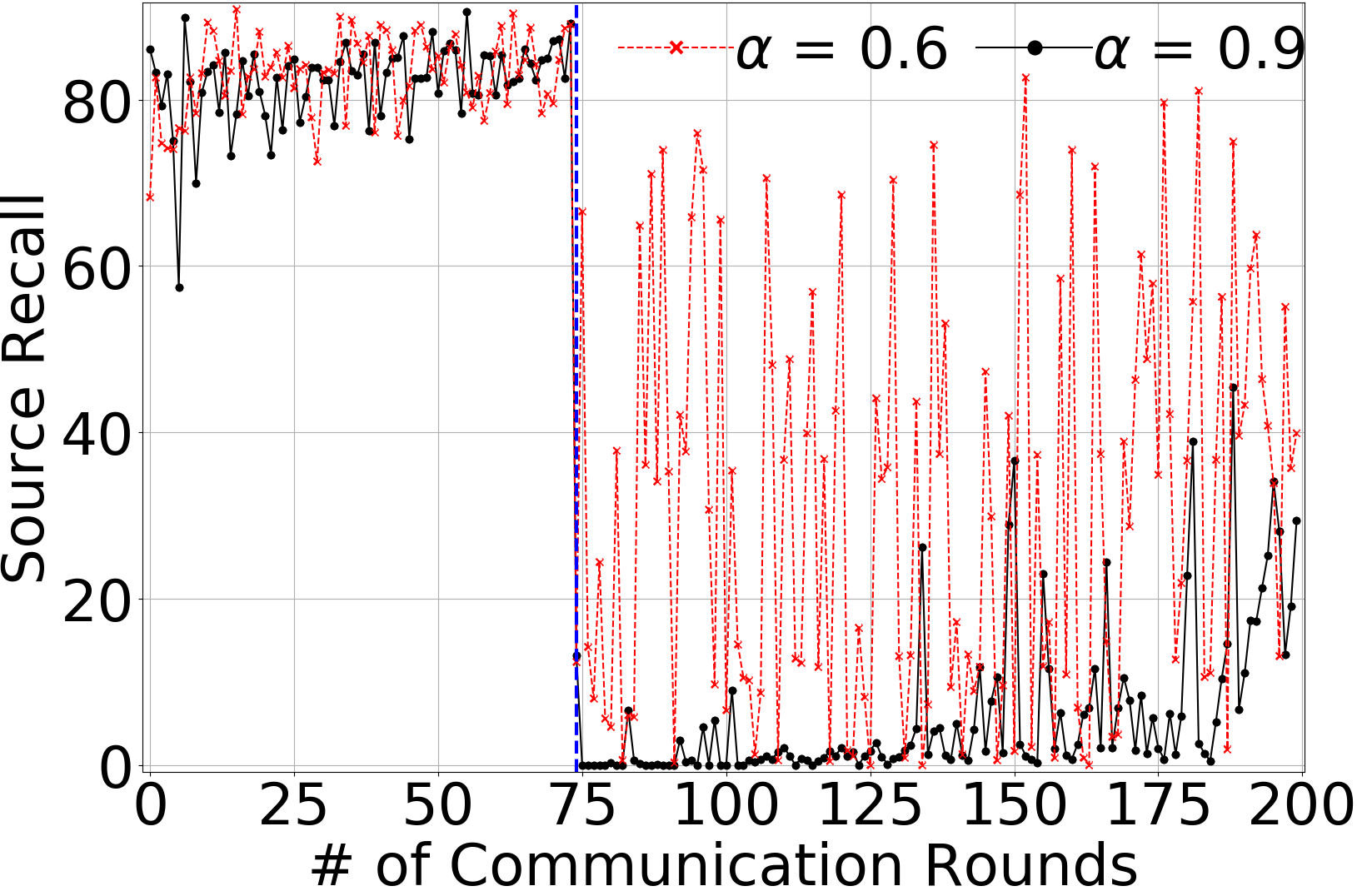}
    \vspace{-3pt}
    \caption{Fashion-MNIST}
  \end{subfigure}
  \vspace{-3pt}
  \caption{Source class recall by round when malicious participants' availability is close to that of honest participants ($\alpha=0.6$) vs 
  significantly increased ($\alpha=0.9$). The blue line indicates the round in which attack starts.
  }
  \label{fig:3B}
  \vspace{-18pt}
\end{figure}

To more acutely demonstrate the impact of $\alpha$, Figure~\ref{fig:3B} reports source class recall by round when $\alpha=0.6$ and $\alpha=0.9$ for both the CIFAR-10 and Fashion-MNIST datasets. In both datasets, when malicious participants are available more frequently, the source class recall is effectively shifted lower in the graph, i.e., source class recall values with $\alpha=0.9$ are often much smaller than those with $\alpha=0.6$. We note that the high round-by-round variance in both graphs is due to the probabilistic variability in number of malicious participants in individual training rounds. When fewer malicious participants are selected in one training round relative to the previous round, source recall increases. When more malicious participants are selected in an individual round relative to the previous round, source recall falls. 

\begin{figure}[!t]
\centering
  \begin{subfigure}{0.45\textwidth}
    \centering\includegraphics[width=0.9\textwidth,]{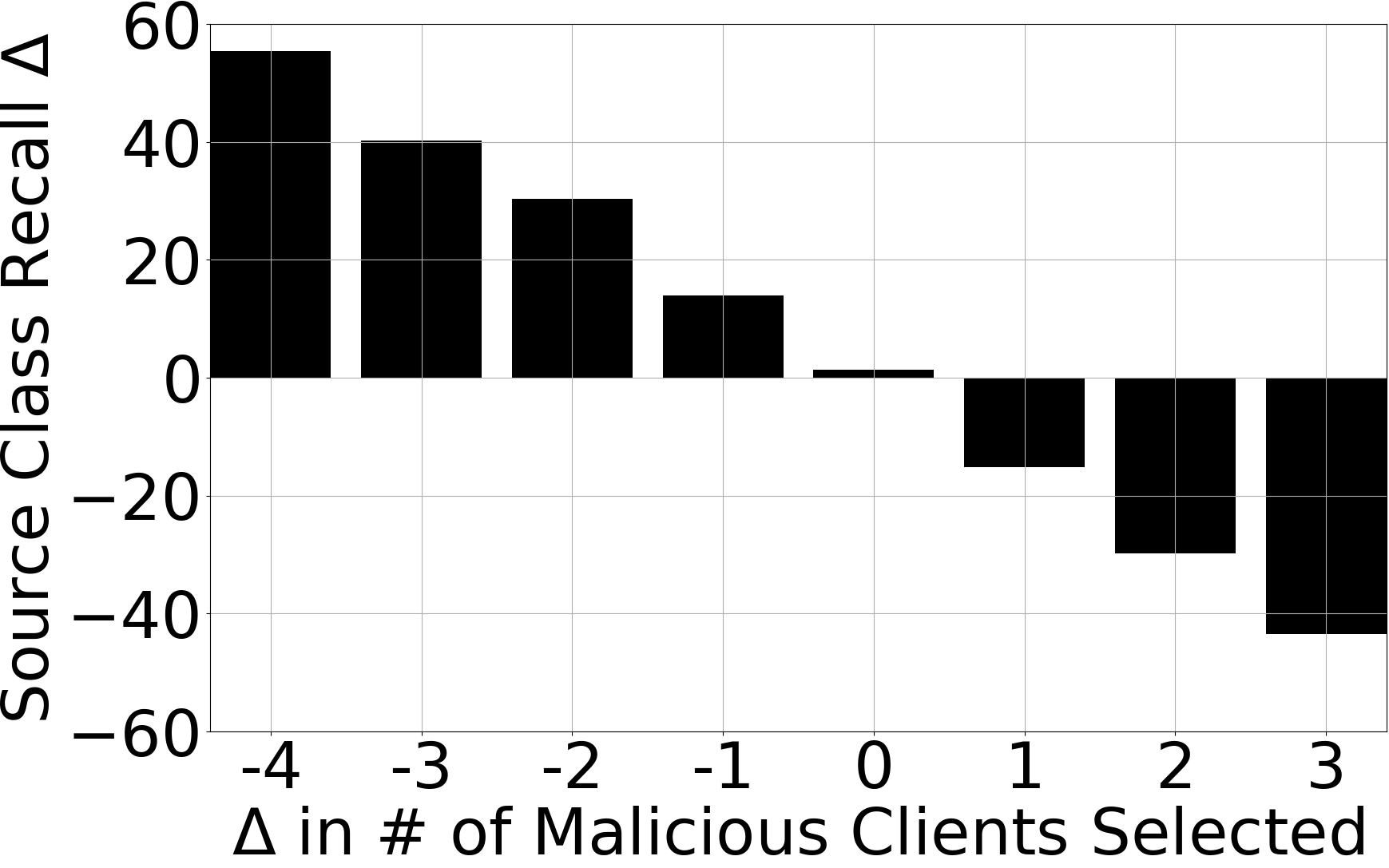}
    \caption{CIFAR-10}
  \end{subfigure}
  \hspace{6pt}
  \begin{subfigure}{0.45\textwidth}
    \centering\includegraphics[width=0.9\textwidth]{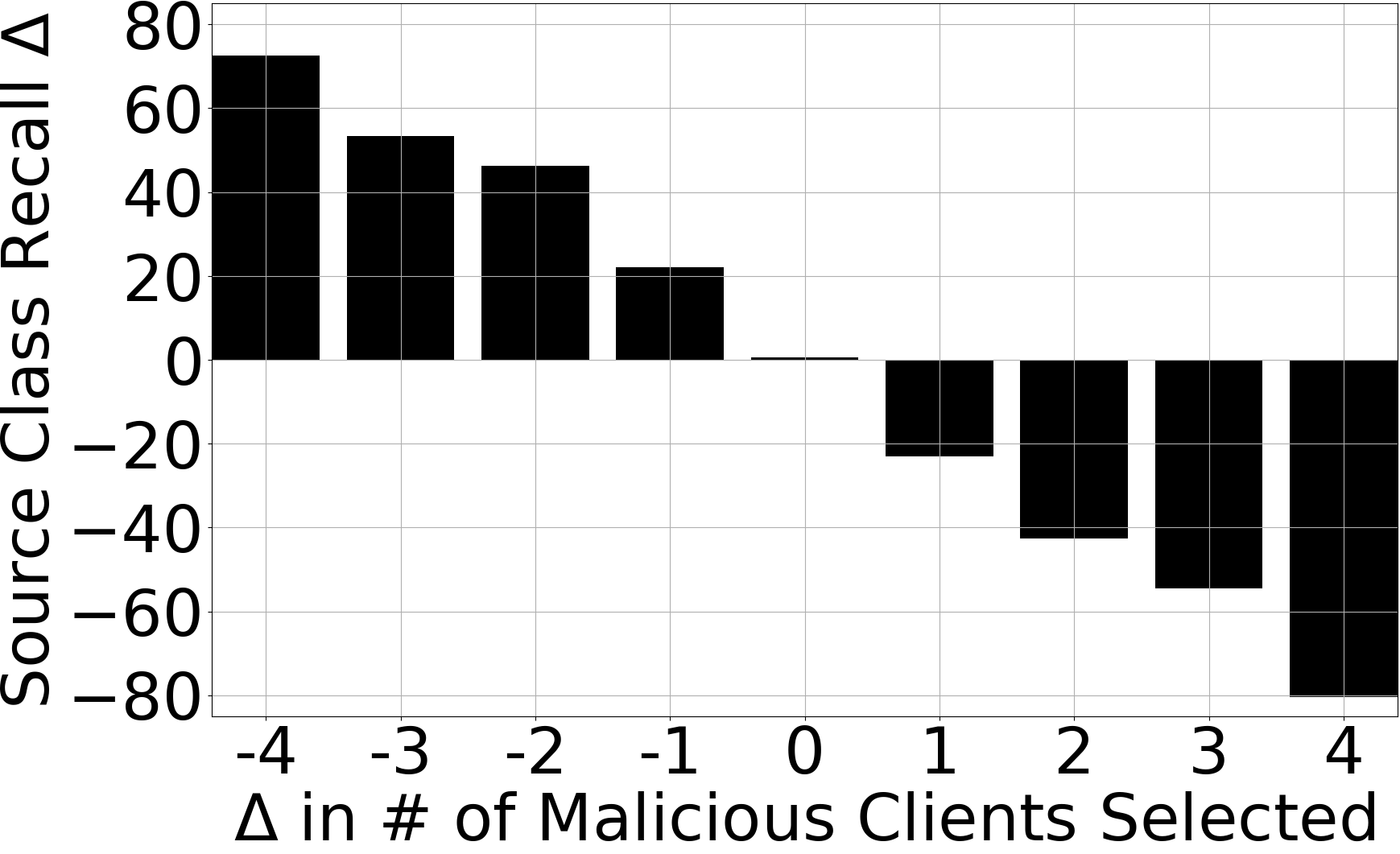}
    \caption{Fashion-MNIST}
  \end{subfigure}
  \caption{Relationship between change in source class recall in consecutive rounds versus change in number of malicious participants in consecutive rounds. Specifically, $\forall r \geq 75$ the y-axis represents $(c_{src}^{recall}$ @ round $r)$ - $(c_{src}^{recall}$ @ round $r-1)$ while the x-axis represents (\# of malicious $\in \mathcal{P}_r$) - (\# of malicious $\in \mathcal{P}_{r-1}$).} 
  \label{fig:3C}
  \vspace{-16pt}
\end{figure}

We further explore and illustrate our last remark with respect to the impact of malicious parties' participation in consecutive rounds in Figure~\ref{fig:3C}. In this figure, the x-axis represents the change in the number of malicious clients participating in consecutive rounds, i.e., (\# of malicious $\in \mathcal{P}_r$) -- (\# of malicious $\in \mathcal{P}_{r-1}$). The y-axis represents the change in source class recall between these consecutive rounds, i.e., $(c_{src}^{recall}$ @ round $r)$ -- $(c_{src}^{recall}$ @ round $r-1)$. The reported results are then averaged across multiple runs of FL and all cases in which each participation difference was observed. The results confirm our intuition that, when $\mathcal{P}_r$ contains more malicious participants than $\mathcal{P}_{r-1}$, there is a substantial drop in source class recall. For large differences (such as +3 or +4), the drop could be as high as 40\% or 60\%. In contrast, when $\mathcal{P}_r$ contains fewer malicious participants than $\mathcal{P}_{r-1}$, there is a substantial increase in source class recall, which can be as high as 60\% or 40\% when the difference is -4 or -3. Altogether, this demonstrates the possibility that the DNN could recover significantly even in few rounds of FL training, if a large enough decrease in malicious participation could be achieved.

\vspace{-12pt}
\section{Defending Against Label Flipping Attacks} \label{sec:defense}
\vspace{-10pt}
Given a highly effective adversary, how can a FL system defend against the label flipping attacks discussed thus far? To that end, we propose a defense which enables the aggregator to identify malicious participants. 

\vspace{-18pt}

\begin{algorithm}[ht]
\DontPrintSemicolon 
\SetAlgoLined
\SetKwProg{Def}{def}{:}{}
\SetKwFunction{evalUpds}{evaluate\_updates}
\SetKwFunction{trainDNN}{train\_DNN}
\SetKwFunction{standardize}{standardize}
\SetKwFunction{PCA}{PCA}
\SetKwFunction{plot}{plot}
\Def{\evalUpds{$\mathcal{R}:$ set of vulnerable train rounds, $\mathcal{P}:$ participant set}}{
    $\mathcal{U} = \emptyset$\;
    \For{$r \in \mathcal{R}$}{
        $\mathcal{P}_r \leftarrow$ participants $\in \mathcal{P}$ queried in training round $r$\;
        $\theta_{r-1} \leftarrow$ global model parameters after training round $r-1$\;
        \For{$P_i \in \mathcal{P}_r$}{
            $\theta_{r, i} \leftarrow$ updated parameters after \trainDNN{$\theta_{r-1}$, $D_i$}\;
            $\theta_{\Delta, i} \leftarrow \theta_{r, i} - \theta_r$\;
            $\theta^{src}_{\Delta,i} \leftarrow$ parameters $\in \theta_{\Delta,i}$ connected to source class output node\;
            Add $\theta^{src}_{\Delta,i}$ to $\mathcal{U}$\;
        }
    }
    $\mathcal{U}' \leftarrow$ \standardize{$\mathcal{U}$}\;
    $\mathcal{U}'' \leftarrow$ \PCA{$\mathcal{U'}$, components=2}\;
    \plot{$\mathcal{U}''$}\;
}
\caption{Identifying Malicious Model Updates in FL}
\label{algo:4a}
\end{algorithm}
\vspace{-18pt}

After identifying malicious participants, the aggregator may blacklist them or ignore their updates $\theta_{r,i}$ in future rounds. We showed in Sections \ref{subsec:timing} and \ref{sec:part-availability} that high-utility model convergence can be eventually achieved after eliminating malicious participation. The feasibility of such a recovery from early round attacks supports use of the proposed identification approach as a defense strategy.

Our defense is based on the following insight: The parameter updates sent from malicious participants have unique characteristics compared to honest participants' updates for a subset of the parameter space. However, since DNNs have many parameters (i.e., $\theta_{r,i}$ is extremely high dimensional) it is non-trivial to analyze parameter updates by hand. Thus, we propose an automated strategy for identifying the relevant parameter subset and for studying participant updates using dimensionality reduction (PCA).

\begin{figure}
\vspace{-10pt}
  \begin{subfigure}{0.24\textwidth}
    \centering\includegraphics[width=\textwidth]{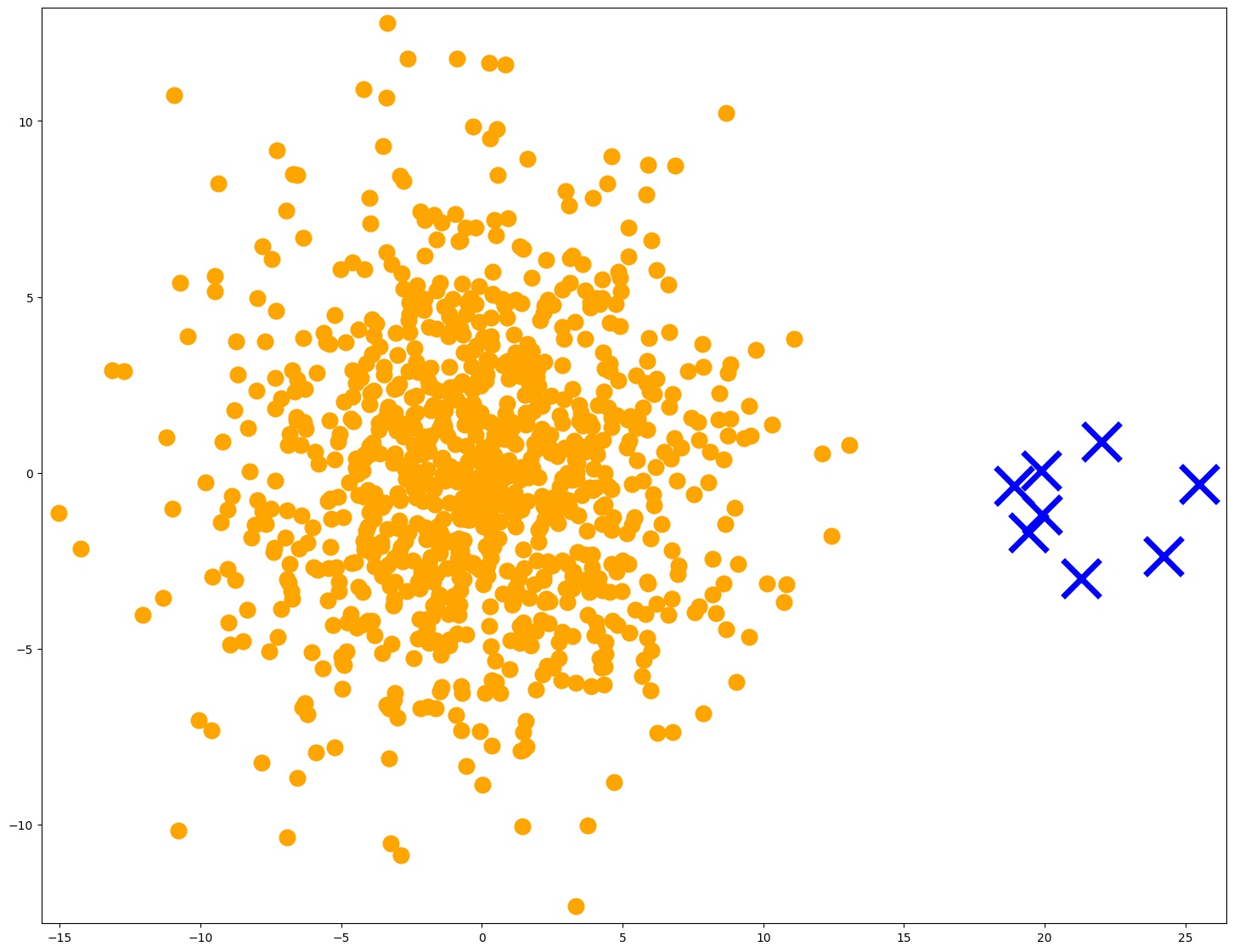}
    \caption{\scriptsize{CIFAR-10 $m$=2\%}}\label{fig:4A-cifar2}
  \end{subfigure}
  \hspace{0.1mm}
  \begin{subfigure}{0.24\textwidth}
    \centering\includegraphics[width=\textwidth]{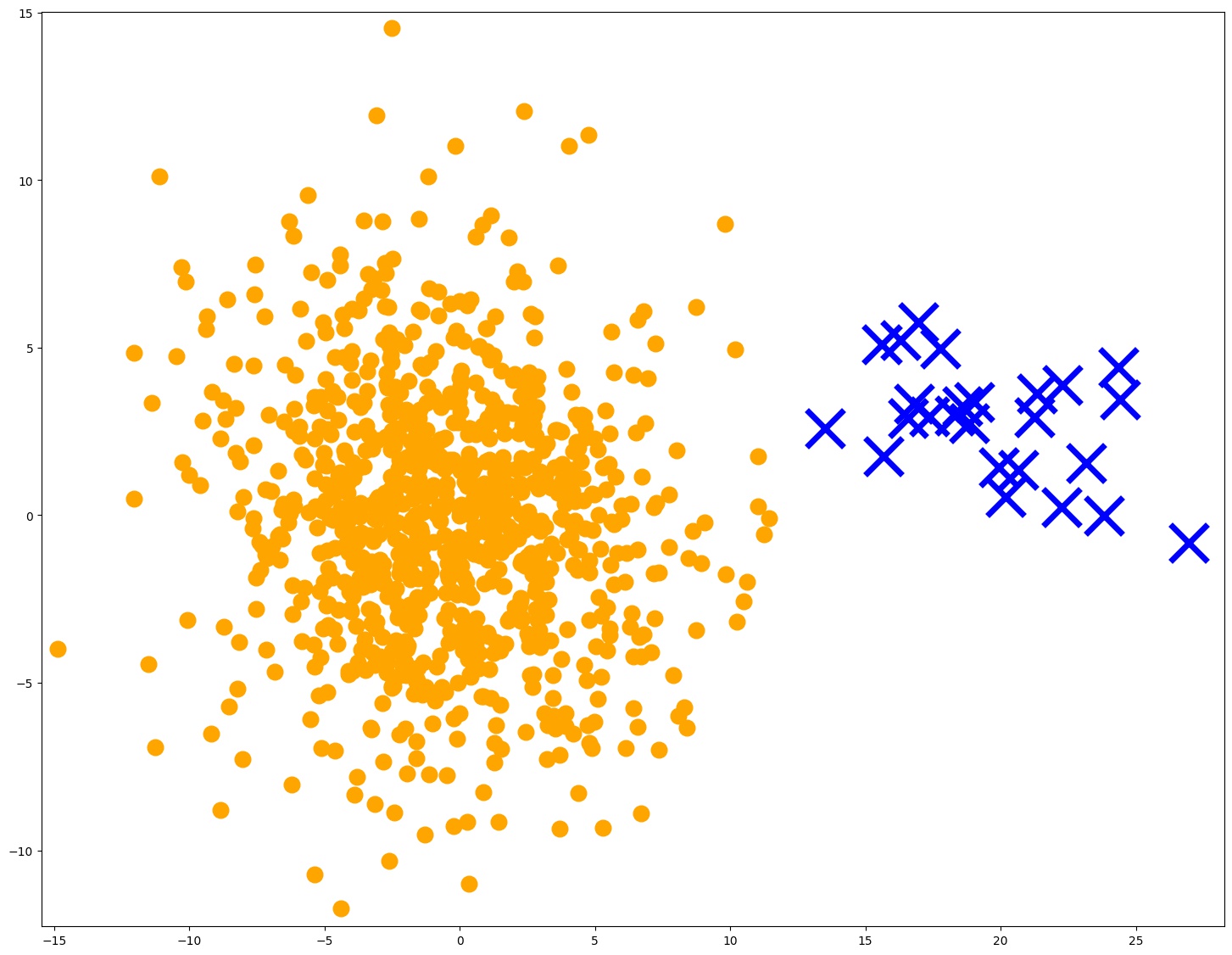}
    \caption{\scriptsize{CIFAR-10 $m$=4\%}}\label{fig:4A-cifar4}
  \end{subfigure}
  \hspace{0.1mm}
  \begin{subfigure}{0.24\textwidth}
    \centering\includegraphics[width=\textwidth]{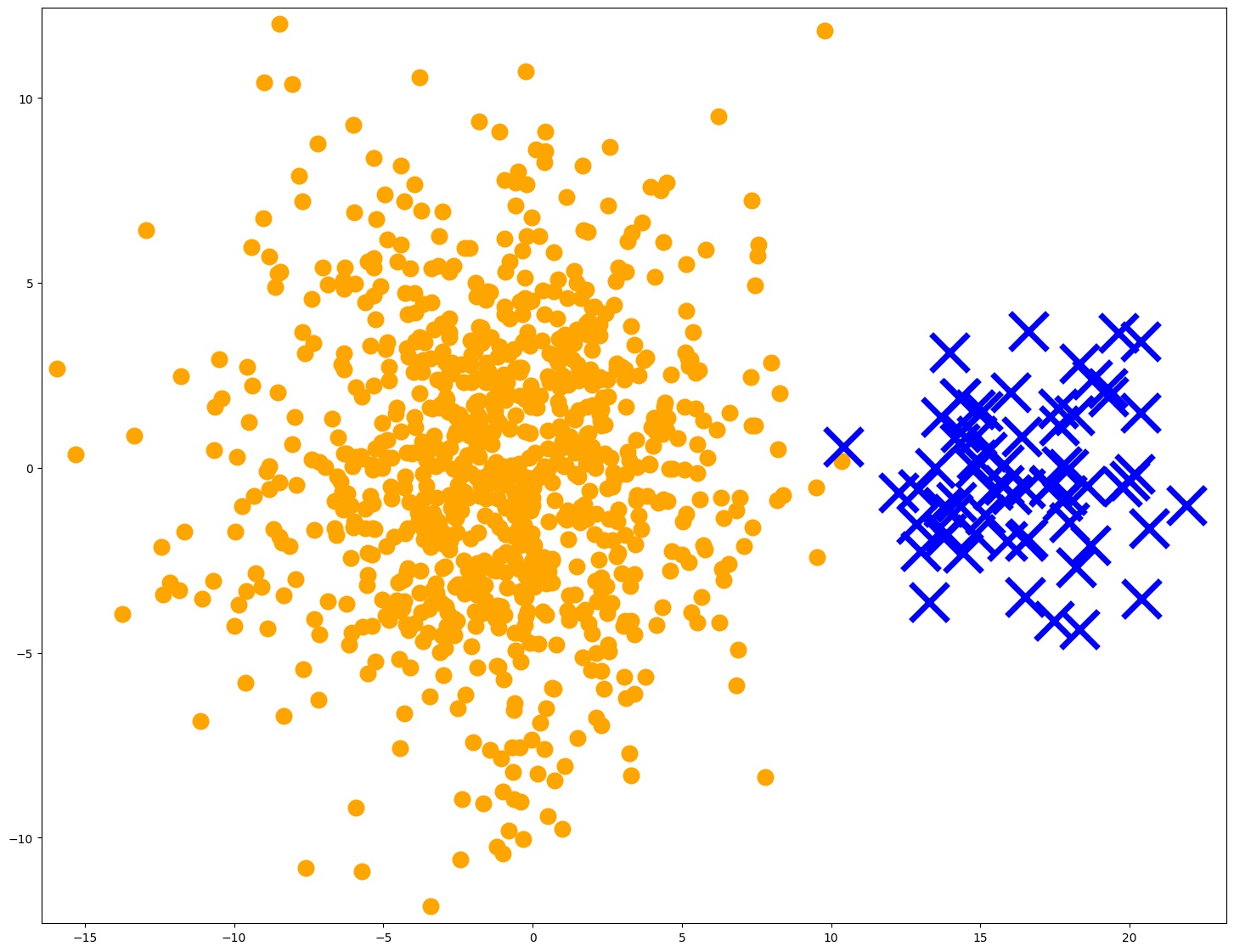}
    \caption{\scriptsize{CIFAR-10 $m$=10\%}}\label{fig:4A-cifar10}
  \end{subfigure}
  \hspace{0.1mm}
  \begin{subfigure}{0.24\textwidth}
    \centering\includegraphics[width=\textwidth]{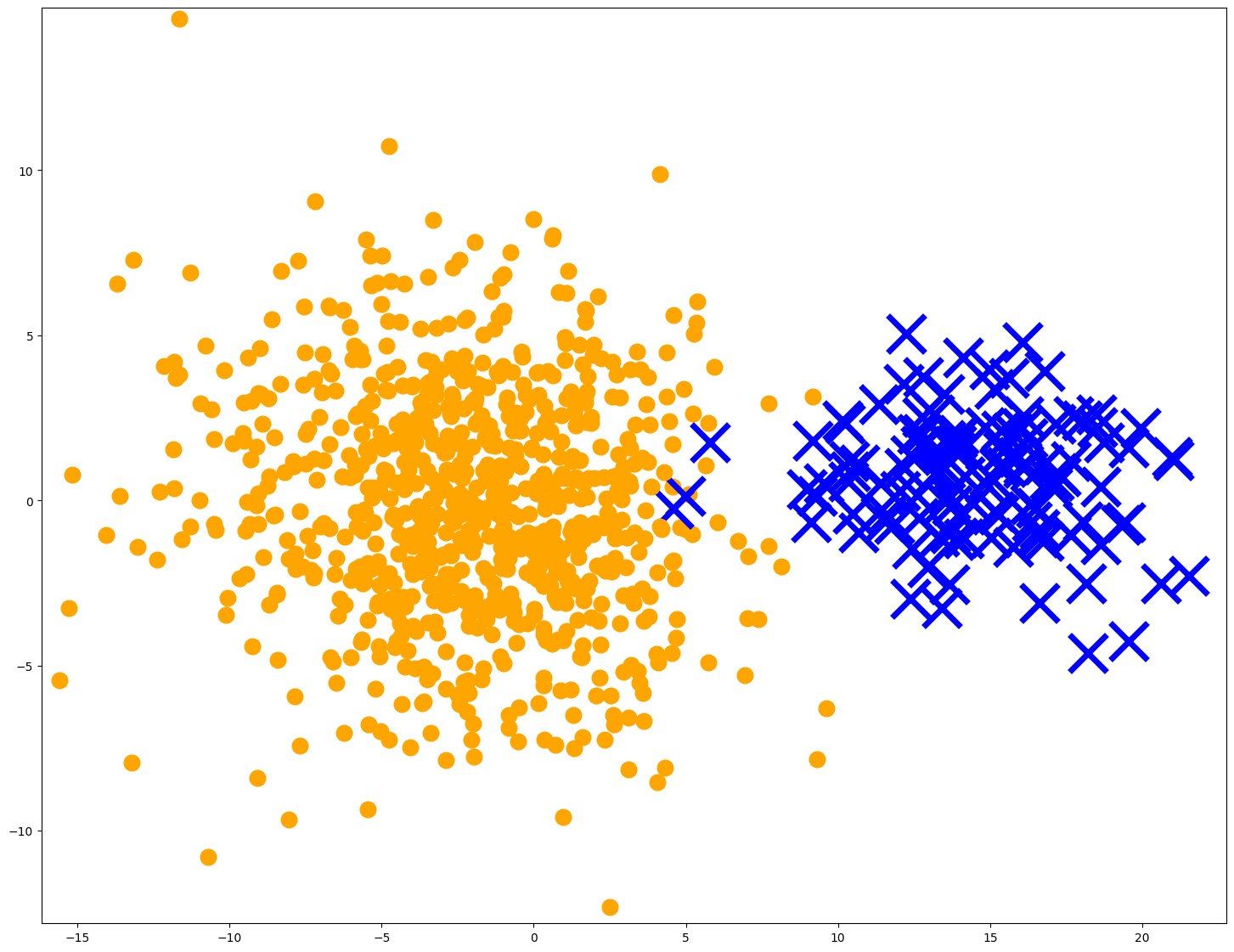}
    \caption{\scriptsize{CIFAR-10 $m$=20\%}}\label{fig:4A-cifar20}
  \end{subfigure}\newline
  \begin{subfigure}{0.24\textwidth}
    \centering\includegraphics[width=\textwidth]{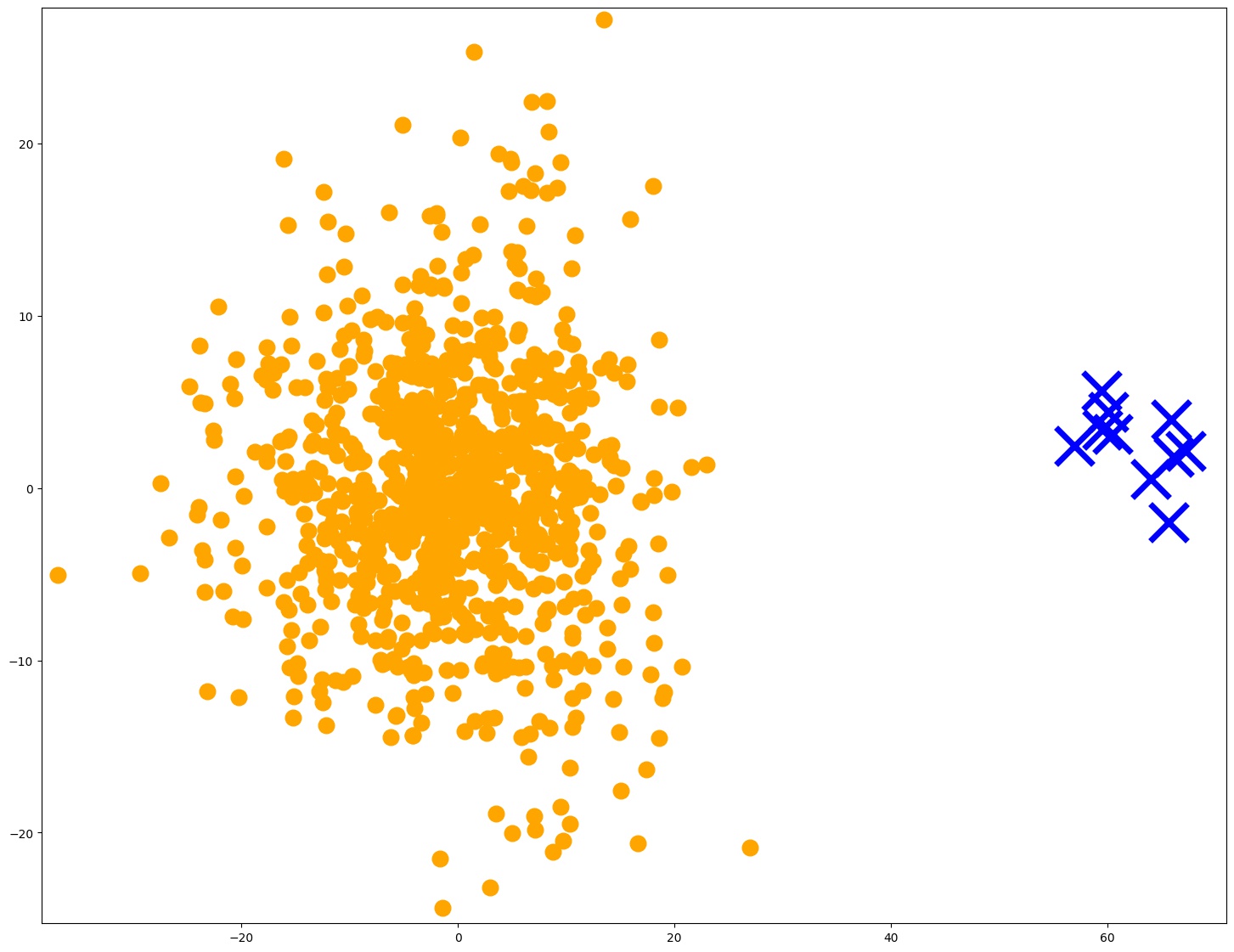}
    \caption{\scriptsize{F-MNIST $m$=2\%}}\label{fig:4A-fmnist2}
  \end{subfigure}
  \hspace{0.1mm}
  \begin{subfigure}{0.24\textwidth}
    \centering\includegraphics[width=\textwidth]{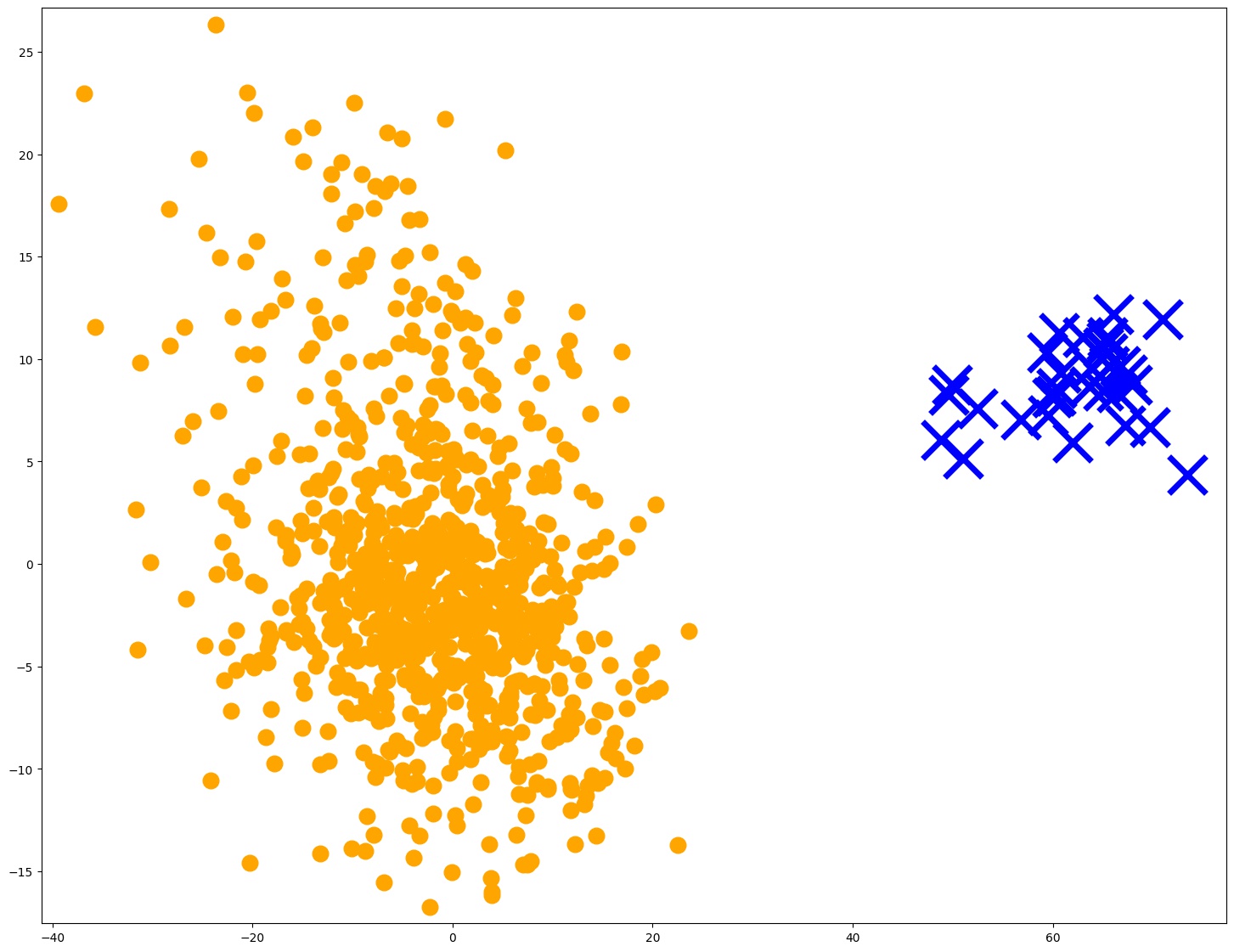}
    \caption{\scriptsize{F-MNIST $m$=4\%}}\label{fig:4A-fmnist4}
  \end{subfigure}
  \hspace{0.1mm}
  \begin{subfigure}{0.24\textwidth}
    \centering\includegraphics[width=\textwidth]{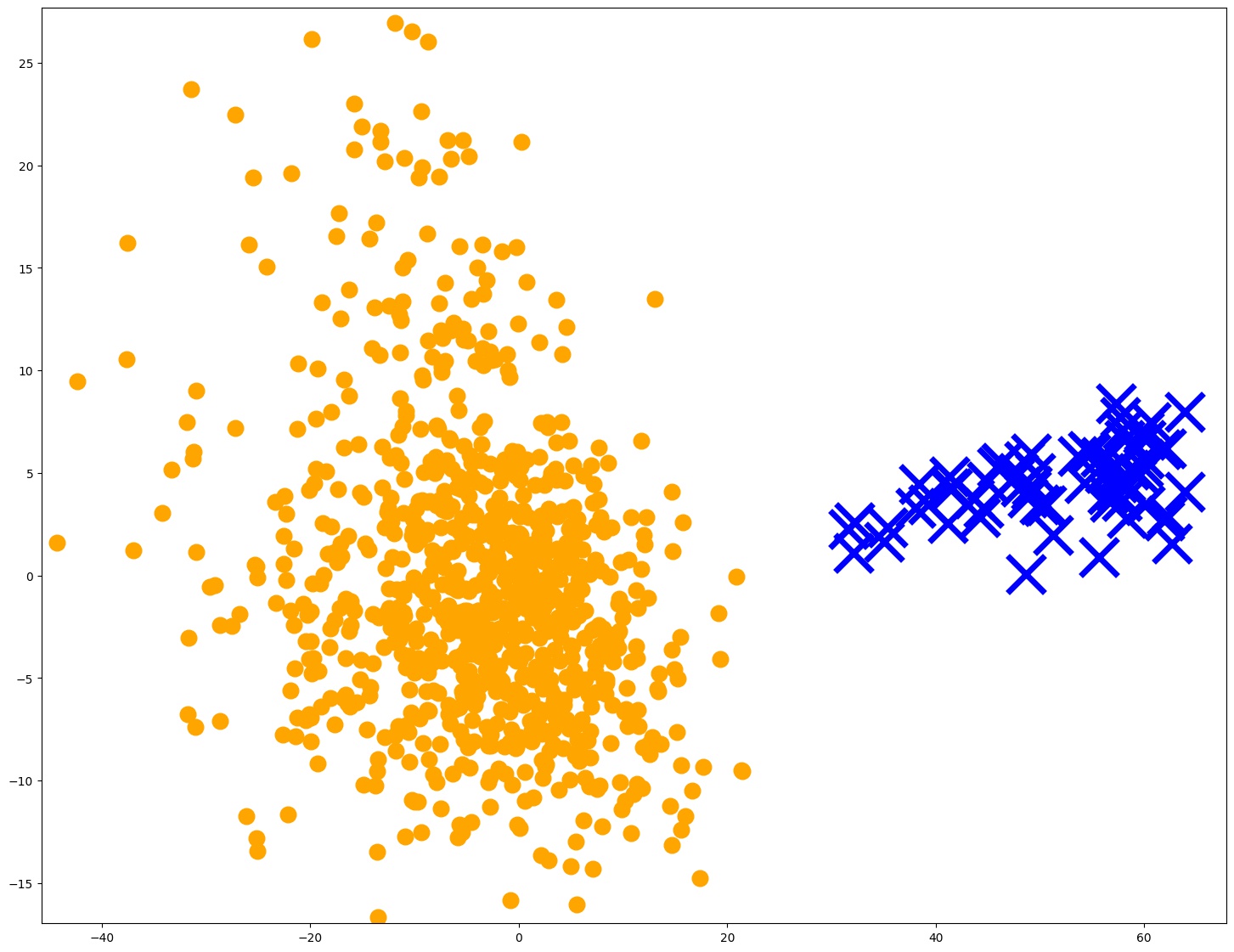}
    \caption{\scriptsize{F-MNIST $m=$10\%}}\label{fig:4A-fmnist10}
  \end{subfigure}
  \hspace{0.1mm}
  \begin{subfigure}{0.24\textwidth}
    \centering\includegraphics[width=\textwidth]{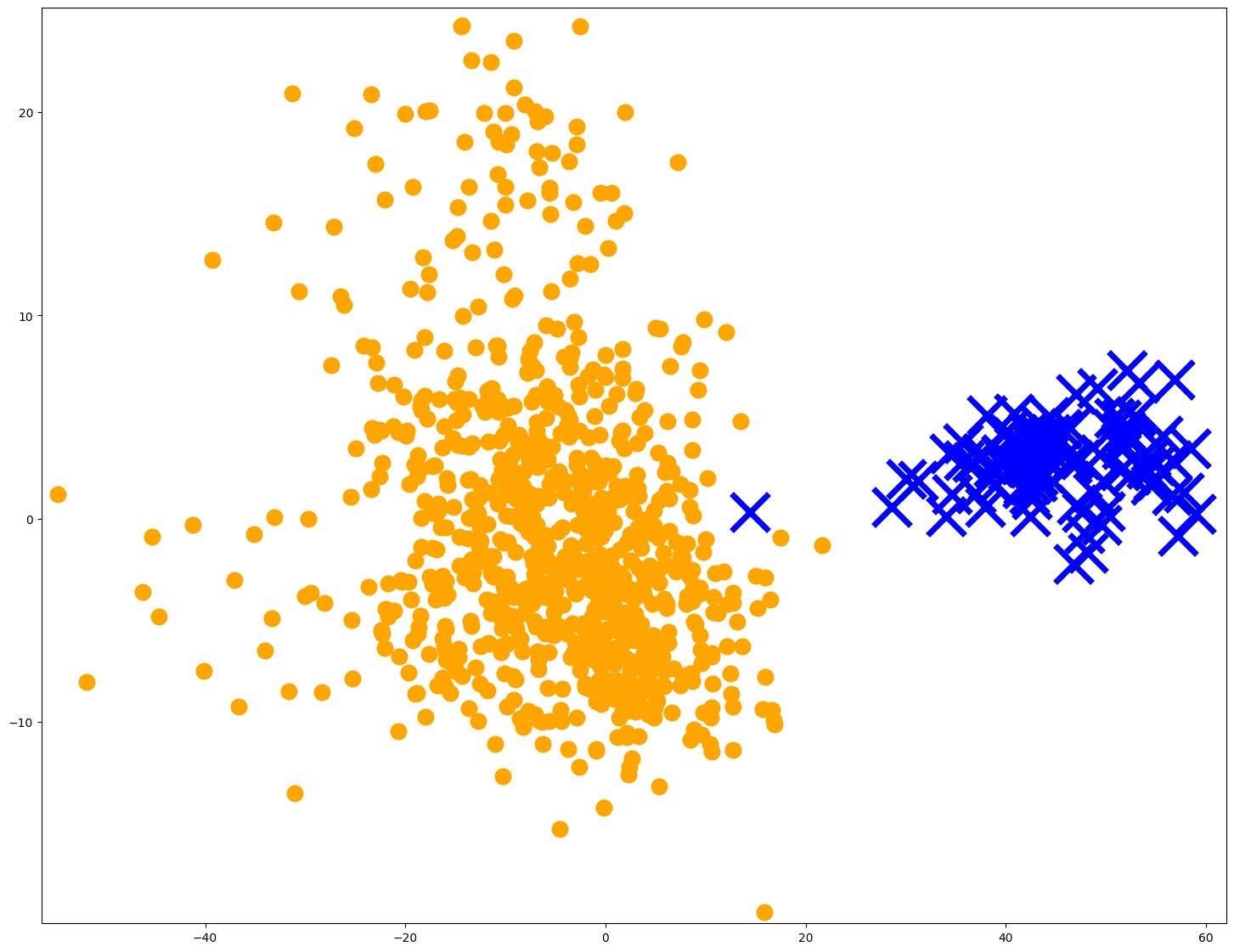}
    \caption{\scriptsize{F-MNIST $m$=20\%}}\label{fig:4A-fmnist20}
  \end{subfigure}
  \caption{PCA plots with 2 components demonstrating the ability of Algorithm \ref{algo:4a} to identify updates originating from a malicious versus honest participant. Plots represent relevant gradients collected from all training rounds $r > 10$. Blue Xs represent gradients from malicious participants while yellow Os represent gradients from honest participants.}
  \label{fig:4A}
  \vspace{-18pt}
\end{figure}

The description of our defense strategy is given in Algorithm \ref{algo:4a}. Let $\mathcal{R}$ denote the set of vulnerable FL training rounds and $c_{src}$ be the class that is suspected to be the source class of a poisoning attack. We note that if $c_{src}$ is unknown, the aggregator can defend against potential attacks such that $c_{src}=c$ $\forall c\in\mathcal{C}$. We also note that for a given $c_{src}$, Algorithm~\ref{algo:4a} considers label flipping for all possible $c_{target}$. An aggregator therefore will conduct $|\mathcal{C}|$ independent iterations of Algorithm~\ref{algo:4a}, which can be conducted in parallel. For each round $r \in \mathcal{R}$ and participant $P_i \in \mathcal{P}_r$, the aggregator computes the delta in participant's model update compared to the global model, i.e., $\theta_{\Delta, i} \leftarrow \theta_{r, i} - \theta_r$. Recall from Section~\ref{subsec:fl} that a predicted probability for any given class $c$ is computed by a specific node $n_c$ in the final layer DNN architecture. Given the aggregator's goal of defending against the label flipping attack from $c_{src}$, only the subset of the parameters in $\theta_{\Delta, i}$ corresponding to $n_{c_{src}}$ is extracted. The outcome of the extraction is denoted by $\theta^{src}_{\Delta,i}$ and added to a global list $\mathcal{U}$ built by the aggregator. After $\mathcal{U}$ is constructed across multiple rounds and participant deltas, it is standardized by removing the mean and scaling to unit variance. The standardized list $\mathcal{U}'$ is fed into Principal Component Analysis (PCA), which is a popular ML technique used for dimensionality reduction and pattern visualization. For ease of visualization, we use and plot results with two dimensions (two components). 

In Figure \ref{fig:4A}, we show the results of Algorithm \ref{algo:4a} on CIFAR-10 and Fashion-MNIST across varying malicious participation rate $m$, with $\mathcal{R} = [10, 200]$. Even in scenarios with low $m$, as is shown in Figures~\ref{fig:4A-cifar2} and \ref{fig:4A-fmnist2}, our defense is capable of differentiating between malicious and honest participants. In all graphs, the PCA outcome shows that malicious participants' updates belong to a visibly different cluster compared to honest participants' updates which form their own cluster. Another interesting observation is that our defense does not suffer from the ``gradient drift" problem. Gradient drift is a potential challenge in designing a robust defense, since changes in model updates may be caused both by actual DNN learning and convergence (which is desirable) or malicious poisoning attempt (which our defense is trying to identify and prevent). Our results show that, even though the defense is tested with a long period of rounds (190 training rounds since $\mathcal{R} = [10, 200]$), it remains capable of separating malicious and honest participants, demonstrating its robustness to gradient drift.

A FL system aggregator can therefore effectively identify malicious participants, and consequently restrict their participation in mobile training, by conducting such gradient clustering prior to aggregating parameter updates at each round. Clustering model gradients for malicious participant identification presents a strong defense as it does not require access to any public validation dataset, as is required in~\cite{baracaldo2017mitigating}, which is not necessarily possible to acquire.

\vspace{-12pt}
\section{Related Work}\label{sec:related_work}
\vspace{-10pt}
Poisoning attacks are highly relevant in domains such as spam filtering \cite{google-url,nelson2008exploiting}, malware and network anomaly detection \cite{chen2018automated,maiorca2019towards,rubinstein2009antidote}, disease diagnosis \cite{mozaffari2014systematic}, computer vision \cite{papernot2018sok}, and recommender systems \cite{fang2018poisoning,yang2017fake}. Several poisoning attacks were developed for popular ML models including SVM \cite{biggio2012poisoning,demontis2019adversarial,steinhardt2017certified,suciu2018does,xiao2012adversarial,xiao2015support}, regression~\cite{jagielski2018manipulating}, dimensionality reduction \cite{xiao2015feature}, linear classifiers \cite{demontis2019adversarial,liu2017robust,zhao2017efficient}, unsupervised learning \cite{biggio2013data}, and more recently, neural networks \cite{demontis2019adversarial,munoz2017towards,shafahi2018poison,suciu2018does,yang2017generative,zhu2019transferable}. However, most of the existing work is concerned with poisoning ML models in the traditional setting where training data is first collected by a centralized party. In contrast, our work studies poisoning attacks in the context of FL. As a result, many of the poisoning attacks and defenses that were designed for traditional ML are not suitable to FL. For example, attacks that rely on crafting optimal poison instances by observing the training data distribution are inapplicable since the malicious FL participant may only access and modify the training data s/he holds. Similarly, server-side defenses that rely on filtering and eliminating poison instances through anomaly detection or k-NN \cite{paudice2018detection,paudice2018label} are inapplicable to FL since the server only observes parameter updates from FL participants, not their individual instances.

The rising popularity of FL has led to the investigation of different attacks in the context of FL, such as backdoor attacks \cite{bagdasaryan2018backdoor,sun2019can}, gradient leakage attacks \cite{hitaj2017deep,melis2019exploiting,zhu2019deep} and membership inference attacks \cite{nasr2019comprehensive,truex2018towards,truex2019demystifying}. Most closely related to our work are poisoning attacks in FL. There are two types of poisoning attacks in FL: \textit{data poisoning} and \textit{model poisoning}. Our work falls under the data poisoning category. In data poisoning, a malicious FL participant manipulates their training data, e.g., by adding poison instances or adversarially changing existing instances \cite{fung2018mitigating,shen2016auror}. The local learning process is otherwise not modified. In model poisoning, the malicious FL participant modifies its learning process in order to create adversarial gradients and parameter updates. \cite{bhagoji2019analyzing} and \cite{fang2019local} demonstrated the possibility of causing high model error rates through targeted and untargeted model poisoning attacks. While model poisoning is also effective, data poisoning may be preferable or more convenient in certain scenarios, since it does not require adversarial tampering of model learning software on participant devices, it is efficient, and it allows for non-expert poisoning participants.

Finally, FL poisoning attacks have connections to the concept of \textit{Byzantine threats}, in which one or more participants in a distributed system fail or misbehave. In FL, Byzantine behavior was shown to lead to sub-optimal models or non-convergence \cite{blanchard2017machine,kairouz2019advances}. This has spurred a line of work on Byzantine-resilient aggregation for distributed learning, such as Krum \cite{blanchard2017machine}, Bulyan \cite{mhamdi2018hidden}, trimmed mean, and coordinate-wise median \cite{yin2018byzantine}. While model poisoning may remain successful despite Byzantine-resilient aggregation \cite{bhagoji2019analyzing,fang2019local,kairouz2019advances}, it is unclear whether optimal data poisoning attacks can be found to circumvent an individual Byzantine-resilient scheme, or whether one data poisoning attack may circumvent multiple Byzantine-resilient schemes. We plan to investigate these issues in future work.

\vspace{-12pt}
\section{Conclusion}\label{sec:conclusion}
\vspace{-10pt}
In this paper we studied data poisoning attacks against FL systems. We demonstrated that FL systems are vulnerable to label flipping poisoning attacks and 
that these attacks can significantly negatively impact the global model. We also showed that the negative impact on the global model increases as the proportion of malicious participants increases, and that it is possible to achieve targeted poisoning impact
. Further, we demonstrated that adversaries can enhance 
attack effectiveness by increasing the availability of malicious participants in later 
rounds. Finally, we proposed a defense which helps an FL aggregator separate malicious from honest participants. We showed that our defense is capable of identifying malicious participants and it is 
robust to gradient drift.

As poisoning attacks against 
FL 
systems continue to emerge as important research topics in the security and ML communities \cite{fang2019local,bhagoji2019analyzing,nguyen2020poisoning,zhao2020shielding,khazbak2020mlguard}, we plan to continue our work in several ways. First, we will study the impacts of the attack and defense on diverse FL scenarios differing in terms of data size, distribution among FL participants (iid vs non-iid), data type, total number of instances available per class, 
etc
. Second, we will study more complex adversarial behaviors such as each malicious 
participant changing the labels of only a small portion of source samples or using more sophisticated poisoning strategies to avoid being detected
. Third, while we designed and tested our defense against the label flipping attack, we hypothesize the defense will be useful against model poisoning attacks since malicious participants’ gradients are often dissimilar 
to 
those of 
honest participants. Since our defense identifies dissimilar or anomalous gradients
, we expect the defense to be effective against other types of FL attacks that cause dissimilar or anomalous gradients. In future work, we will study the applicability of our defense against such other FL attacks including model poisoning, untargeted poisoning, and backdoor attacks.


\noindent\textbf{Acknowledgements}. This research is partially sponsored by NSF CISE SaTC 1564097. The second author acknowledges an IBM PhD Fellowship Award and the support from the Enterprise AI, Systems \& Solutions division led by Sandeep Gopisetty at IBM Almaden Research Center. Any opinions, findings, and conclusions or recommendations expressed in this material are those of the author(s) and do not necessarily reflect the views of the National Science Foundation or other funding agencies and companies mentioned above.
\vspace{-12pt}

%
%
%
\bibliographystyle{splncs04}
\bibliography{labelflip_fl}

\appendix
\vspace{-16pt}
\section{DNN Architectures and Configuration}\label{appendix:a}
\vspace{-8pt}
All NNs were trained using PyTorch version 1.2.0 with random weight initialization. Training and testing was completed using a NVIDIA 980 Ti GPU-accelerator. When necessary, all CUDA tensors were mapped to CPU tensors before 
exporting to Numpy arrays. Default drivers provided by Ubuntu 19.04 and 
built-in GPU support in PyTorch was used to accelerate training. 
Details can be found in our 
repository: \url{https://github.com/git-disl/DataPoisoning_FL}.

\noindent\textbf{Fashion-MNIST:} We do not conduct 
data 
pre-processing
. We use a Convolutional Neural Network 
with the architecture described in Table \ref{tab:AppendixA2}. In the table, Conv = Convolutional Layer, and Batch Norm = Batch Normalization.

\noindent\textbf{CIFAR-10:} We conduct data pre-processing prior to 
training. 
Data is normalized 
with mean [0.485, 0.456, 0.406] and standard deviation [0.229, 0.224, 0.225]. 
Values reflect mean and standard deviation of the ImageNet dataset~\cite{deng2009imagenet} and are commonplace, even expected when using Torchvision~\cite{marcel2010torchvision} 
models. We additionally perform data augmentation with random horizontal flipping, random cropping with size 32, and default padding. 
Our CNN is 
detailed in Table \ref{tab:AppendixA1}.

\vspace{-12pt}

\begin{table}[!htb]
    \begin{minipage}{.5\textwidth}
        \centering
        \begin{tabular}{|c|c|}
            \hline
            Layer Type & Size \\
            \hline
            Conv + ReLu + Batch Norm & 3x3x32\\
            \hline
            Conv + ReLu + Batch Norm & 3x32x32\\
            \hline
            Max Pooling & 2x2 \\
            \hline
            Conv + ReLu + Batch Norm & 3x32x64\\
            \hline
            Conv + ReLu + Batch Norm & 3x64x64\\
            \hline
            Max Pooling & 2x2 \\
            \hline
            Conv + ReLu + Batch Norm & 3x64x128\\
            \hline
            Conv + ReLu + Batch Norm & 3x128x128\\
            \hline
            Max Pooling & 2x2 \\
            \hline
            Fully Connected & 2048 \\
            \hline
            Fully Connected + Softmax & 128 / 10 \\
            \hline
        \end{tabular}
        \caption{CIFAR-10 CNN.}
        \label{tab:AppendixA1}
    \end{minipage}
    \begin{minipage}{.5\textwidth}
        \centering
        \begin{tabular}{|c|c|}
            \hline
            Layer Type & Size \\
            \hline
            Conv + ReLu + Batch Norm & 5x1x16\\
            \hline
            Max Pooling & 2x2 \\
            \hline
            Conv + ReLu + Batch Norm & 5x16x32\\
            \hline
            Max Pooling & 2x2 \\
            \hline
            Fully Connected & 1568 / 10 \\
            \hline
        \end{tabular}
        \caption{Fashion-MNIST CNN.}
        \label{tab:AppendixA2}
    \end{minipage}
\end{table}

\end{document}